\pdfoutput=1

\documentclass[11pt]{article}

\usepackage{emnlp2021}
\usepackage{times}
\usepackage{latexsym}

\usepackage[T1]{fontenc}

\usepackage[utf8]{inputenc}

\usepackage{microtype}

\usepackage{url}
\usepackage{CJKutf8}
\usepackage{graphicx}
\usepackage{amsmath}
\usepackage{amssymb}
\usepackage{makecell}
\usepackage{multirow}
\usepackage{booktabs}
\usepackage{color}
\usepackage{ulem}
\usepackage{bm}

\title{K-PLUG: Knowledge-injected Pre-trained Language Model for Natural Language Understanding and Generation in E-Commerce}

\author{Song Xu\textsuperscript{\rm 1}, Haoran Li\textsuperscript{\rm 1}, Peng Yuan\textsuperscript{\rm 1}, 
	Yujia Wang\textsuperscript{\rm 2}, Youzheng Wu\textsuperscript{\rm 1},\\ \textbf{Xiaodong He}\textsuperscript{\rm 1}, \textbf{Ying Liu}\textsuperscript{\rm 3}, \textbf{Bowen Zhou}\textsuperscript{\rm 1} \\ \\
	\textsuperscript{\rm 1} JD AI Research\\
	\textsuperscript{\rm 2} University of California, Berkeley\\ 
	\textsuperscript{\rm 3} Renmin University of China\\ 
	\texttt{\{xusong28, lihaoran24, yuanpeng29\}@jd.com}\\
}

\begin{document}\begin{CJK*}{UTF8}{gbsn} 
		
		\maketitle
		
		\begin{abstract}
			Existing pre-trained language models (PLMs) have demonstrated the effectiveness of self-supervised learning for a broad range of natural language processing (NLP) tasks. However, most of them are not explicitly aware of domain-specific knowledge, which is essential for downstream tasks in many domains, such as tasks in e-commerce scenarios.
			In this paper,
			we propose K-PLUG, 
			a knowledge-injected pre-trained language model based on the encoder-decoder transformer that can be transferred to both natural language understanding and generation tasks.
			We verify our method in a diverse range of e-commerce scenarios that require domain-specific knowledge. 
			Specifically, we propose five knowledge-aware self-supervised pre-training objectives to formulate the learning of domain-specific knowledge, including e-commerce domain-specific knowledge-bases, aspects of product entities,  categories of product entities, and unique selling propositions of product entities. 
			K-PLUG achieves new state-of-the-art results on a suite of domain-specific NLP tasks, including product knowledge base completion, abstractive product summarization, and multi-turn dialogue,
			significantly outperforms baselines across the board, which demonstrates that the proposed method effectively learns a diverse set of domain-specific knowledge for both language understanding and generation tasks.
			
		\end{abstract}
		
		\section{Introduction}
		
		Pre-trained language models (PLMs), such as ELMo~\citep{peters-etal-2018-deep}, 
		GPT~\citep{GPT}, BERT~\citep{devlin2018bert},  RoBERTa~\citep{liu2019roberta}, and XLNet~\citep{yang2019xlnet},  have made remarkable breakthroughs in many natural language understanding (NLU) tasks, including text classification, reading comprehension, and natural language inference. 
		These models are trained on large-scale text corpora with self-supervision based on either bi-directional or auto-regressive pre-training.
		Equally promising performances have been achieved in natural language generation (NLG) tasks, such as machine translation and text summarization, by  MASS~\citep{song2019mass}, UniLM~\citep{UNILM},  BART~\citep{BART}, T5~\citep{raffel2019exploring}, PEGASUS~\citep{zhang2019pegasus}, and ProphetNet~\citep{yan2020prophetnet}. In contrast, these approaches adopt Transformer-based sequence-to-sequence models to jointly pre-train for both the encoder and the decoder.
		
		While these PLMs can learn rich semantic patterns from raw text data and thereby enhance downstream NLP applications, many of them do not explicitly model domain-specific knowledge.
		As a result, they may not be as sufficient for capturing human-curated or domain-specific knowledge that is necessary for tasks in a certain domain, such as tasks in e-commerce scenarios.
		In order to overcome this limitation, several recent studies have proposed to enrich PLMs with explicit knowledge, including knowledge base (KB)~\citep{zhang2019ernie,peters2019knowledge,xiong2019pretrained,wang2019kepler,wang2020k}, lexical relation~\citep{lauscher2019informing,wang2020k}, word sense~\citep{levine2019sensebert}, part-of-speech tag~\citep{ke2019sentilr}, and sentiment polarity~\citep{ke2019sentilr,tian2020skep}. However, these methods only integrate domain-specific knowledge into the encoder, and the decoding process in many NLG tasks benefits little from these knowledge.
		
		To mitigate this problem,  we propose a \textbf{K}nowledge-injected \textbf{P}re-trained \textbf{L}anguage model that is suitable  for both Natural Language \textbf{U}nderstanding and \textbf{G}eneration (\textbf{K-PLUG}).
		Different from existing knowledge-injected PLMs, K-PLUG integrates knowledge into pre-training for both the encoder and the decoder, and thus K-PLUG can be adopted to both downstream knowledge-driven NLU and NLG tasks.
		We verify the performance of the proposed method in various e-commerce scenarios.
		In the proposed K-PLUG, we formulate the learning of four types of domain-specific knowledge: e-commerce domain-specific knowledge-bases, aspects of product entities,  categories of product entities, and unique selling propositions (USPs)~\citep{reeves2017reality} of product entities.
		Specifically, e-commerce KB stores standardized product attribute information,
		product aspects are features that play a crucial role in understanding product information,
		product categories are the backbones for constructing taxonomies for organization,
		and USPs are the essence of what differentiates a product from its competitors.
		K-PLUG learns these types of knowledge into a unified PLM, enhancing performances for various language understanding and generation tasks.
		
		To effectively learn these four types of valuable domain-specific  knowledge in K-PLUG, we proposed five new pre-training objectives: knowledge-aware masked language model (KMLM), knowledge-aware masked sequence-to-sequence (KMS2S), product entity aspect boundary detection (PEABD), product entity category classification (PECC), and product entity aspect summary generation (PEASG). 
		Among these objectives, KMLM and KMS2S learn to predict 
		the masked single and multiple tokens, respectively, that are associated with domain-specific knowledge rather than general information;
		PEABD determines the boundaries between descriptions of different product aspects given full product information; PECC identifies the product category that each product belongs to; and PEASG generates a summary for each individual product aspect based on the entire product description.
		
		After pre-training K-PLUG, we fine-tune it on  three domain-specific NLP tasks,  namely, e-commerce knowledge base completion, abstractive product  summarization, and multi-turn  dialogue. The  results show that K-PLUG significantly outperforms comparative models on all these tasks.
		
		Our main contributions can be summarized as follows:
		\begin{itemize}
			\item We present K-PLUG that  learns domain-specific knowledge for both the encoder and the decoder in a pre-training language model framework, which benefits both NLG and NLU tasks.
			\item We formulate the learning of four types of domain-specific knowledge in e-commerce scenarios: e-commerce domain-specific knowledge-bases, aspects of product entities,  categories of product entities, and unique selling propositions of product entities, 
			which provide critical information for many applications   in the domain  of e-commerce. Specifically,   five self-supervised objectives are proposed  to learn these four types of knowledge into a unified PLM.
			\item Our proposed model exhibits clear  effectiveness in many  downstream  tasks in the e-commerce scenario, including e-commerce KB completion, abstractive product  summarization, and multi-turn dialogue. 
		\end{itemize}

		\begin{figure*}	
			\centering
			\includegraphics[width=5.5  in]{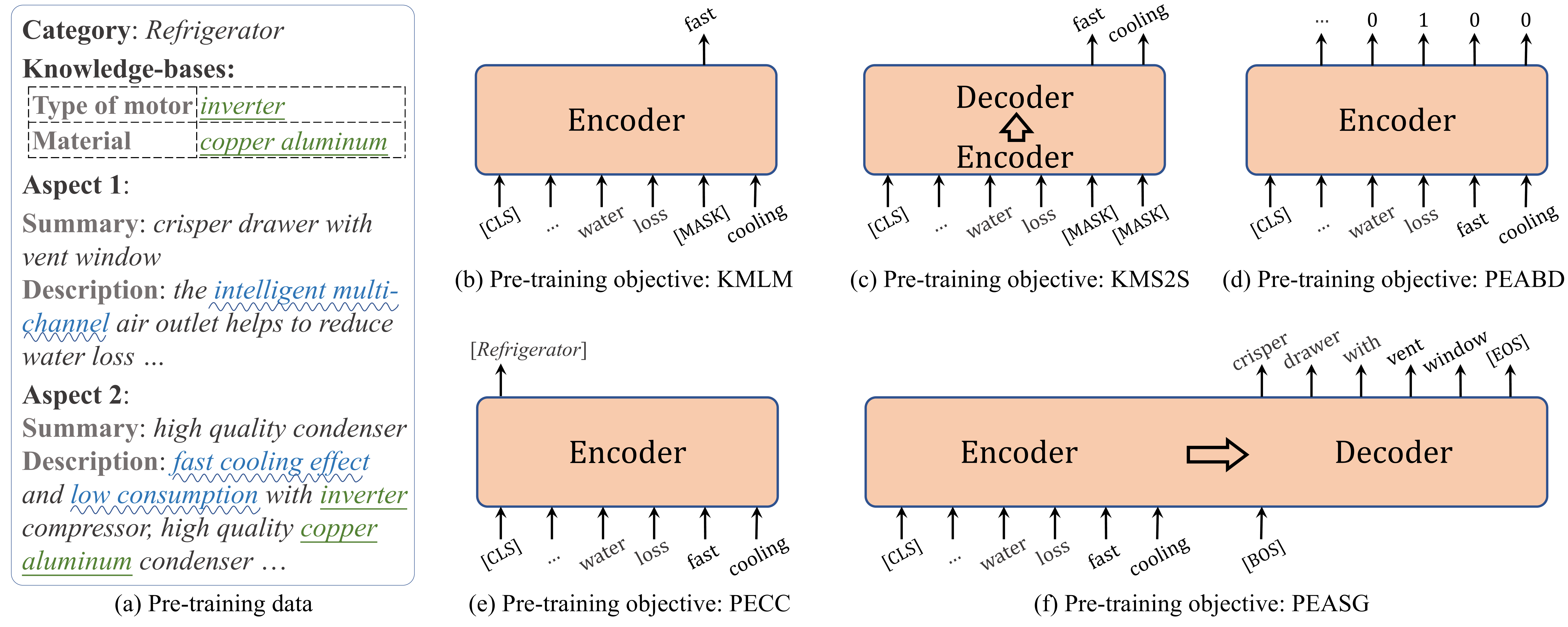}
			\caption{\label{fig}Our pre-training data consists of 25 million textual product descriptions depicting multiple product aspects. We define knowledge  as \textcolor[RGB]{84,130,52}{\underline{\textit{e-commerce knowledge-bases}}}, \textit{aspects of product entities}, \textit{categories of product entities}, and \textcolor[RGB]{46,117,182}{\uwave{\textit{unique selling propositions of product entities}}}. Pre-training objectives include knowledge-aware masked language model (KMLM), knowledge-aware masked sequence-to-sequence (KMS2S), product entity aspect boundary detection (PEABD), product entity category classification (PECC), and product entity aspect summary generation (PEASG).}
		\end{figure*}
		
		\section{Related Work}
		
		\subsection{PLMs in General}
		
		Unsupervised pre-training language model has been successfully applied to many NLP tasks. 
		ELMo~\citep{peters-etal-2018-deep} learns the contextual representations based on a bidirectional LM.
		GPT~\citep{GPT} predicts tokens based on the context on the left-hand side.  BERT~\citep{devlin2018bert}  adopts a bi-directional LM to predict the masked tokens. 
		XLNet~\citep{yang2019xlnet} predicts masked tokens in a permuted order through an autoregressive method.
		MASS~\citep{song2019mass} pre-trains the sequence-to-sequence LM to recover a span of masked tokens. 
		UniLM~\citep{UNILM} combines bidirectional, unidirectional, and sequence-to-sequence LMs. 
		T5~\citep{raffel2019exploring} and BART~\citep{BART} present denoising sequence-to-sequence pre-training.
		PEGASUS~\citep{zhang2019pegasus} pre-trains with gap-sentence generation objective. 
		While human-curated or domain-specific knowledge is essential for downstream knowledge-driven tasks, these methods do not explicitly consider external knowledge like our proposed K-PLUG.

		\subsection{Injecting Knowledge into PLMs}
		Recent work investigates how to incorporate knowledge into PLMs for NLU.
		ERNIE~\citep{sun2019ernie}  enhances language representation with the entity/phrase-level masking.
		ERNIE~\citep{zhang2019ernie} identifies and links entity mentions in texts to their corresponding entities in KB.
		Similar to ERNIE~\citep{zhang2019ernie}, KnowBERT~\citep{peters2019knowledge} injects KBs into 	PLM.
		\citet{xiong2019pretrained} leverages an entity replacement pre-training objective to learn better representations for entities.
		KEPLER~\citep{wang2019kepler} adopts the knowledge embedding objective in the pre-training.
		Besides, SKEP~\citep{tian2020skep}, SenseBERT~\citep{levine2019sensebert}, SentiLR~\citep{ke2019sentilr}, and K-ADAPTER~\citep{wang2020k} propose to integrate sentiment knowledge, word sense, sentiment polarity, and lexical relation into PLM, respectively.
		However, most of these studies are focused on integrating knowledge for language understanding task, work of utilizing domain-specific knowledge for pre-training for language generation tasks are limited. Inspired by these work,  we construct K-PLUG that learns domain-specific knowledge into a PLM for both NLU and NLG tasks.

		\section{Knowledge-injected Pre-training}
		
		In this section, we explain the data used to pre-train K-PLUG, its model architecture, and our pre-training objectives.
		
		\subsection{Data Preparation}
		
		We collect the pre-training data from a mainstream Chinese e-commerce platform\footnote{https://www.jd.com/}, which contains approximately 25 million textual product descriptions and covers 40 product categories. With an average length of 405 tokens, these product descriptions constitute a corpus with a size of 10B Chinese characters. Each product description consists of information on 10.7 product aspects on average, and each product aspect is accompanied with a summary highlighting its prominent features, as shown in Figure~\ref{fig}(a). Additionally, the e-commerce KB and USPs (further explained below) used in our pre-training data are as specified by the e-commerce platform and its online stores. 
		
		\subsection{Model Architecture}
		K-PLUG  adopts the standard sequence-to-sequence Transformer architecture~\citep{vaswani2017attention}, consisting of a 6-layer encoder and a 6-layer decoder as \citet{song2019mass}.
		We set the size of hidden vectors as 768, and the number of self-attention heads as 12. 
		We adopt GELU activation~\citep{hendrycks2016gaussian} as  in GPT~\citep{GPT}.
		We use Adam optimizer~\citep{kingma2014adam} with a learning rate of 5e-4, $\beta_1$ = 0.9, $\beta_2$= 0.98, L2 weight decay of 0.01, learning rate warm-up over the first 10,000 steps and linear decay of the learning rate. The dropout probability is 0.1. The maximum sequence length is set to 512 tokens.
		Pre-training was performed with 4 Telsa V100 GPUs.  The pre-training is done within 10 epochs, which takes around 10 days, and the fine-tuning takes up to 1 day.
		We use the beam search with a beam size of 5 for inference for the NLG tasks.

		\subsection{Knowledge Formulation and Pre-training Objectives}
		
		We formulate the learning of four types of knowledge in a unified PLM: e-commerce KB, aspects of product entities,  categories of product entities, and USPs of product entities.
		Specifically,  \textbf{e-commerce KB}  stores standardized product attribute information, \textit{e.g.}, (\textit{Material}: \textit{Cotton}) and (\textit{Collar Type}: \textit{Pointed Collar}). 
		It provides details about the products~\citep{logan2017multimodal}.
		\textbf{Aspects of product entities} are features of a product, such as the \textit{sound quality} of a stereo speaker, etc.~\citep{li-aaai2020-product}. 
		\textbf{Categories of product entities} such as \textit{Clothing} and \textit{Food} are widely used by e-commerce platforms to organize their products so to present structured offerings to their customers~\citep{luo2020alicoco,dong2020autoknow}
		\textbf{USPs of product entities} are the essence of what differentiates a product from its competitors~\citep{reeves2017reality}. For example, a stereo speaker's USP exhibiting its supreme sound quality could be ``\textit{crystal clear stereo sound}''. An effective USP immediately motivates the purchasing behavior of potential buyers.

		
		
		We propose and evaluate  five novel self-supervised pre-training objectives to learn the above-mentioned four types of knowledge in the K-PLUG model (see Figure~\ref{fig}).
		
		\textbf{Knowledge-aware Masked Language Model (KMLM)}
		
		Inspired by BERT~\citep{devlin2018bert}, we adopt the masked language model (MLM) 
		to train the Transformer encoder as one of our pre-training objectives, which learns to predict the masked tokens in the source sequence (\textit{e.g.}, ``The company is [MASK] at the foot of a hill.'').
		Similar to BERT, we mask 15\% of all tokens in a text sequence; 80\% of the masked tokens are replaced with the [MASK] token, 10\% with a random token, and 10\% left unchanged. Particularly, given an original text sequence ${\bm x} = (x_1,...,x_m,...,x_M)$ with $M$ tokens, 
		a masked sequence  is produced by masking $x_m$  through one of the three ways explained above, \textit{e.g.}, replacing  $x_m$ with [MASK] to create $\widetilde{\bm x} = (x_1,...,\rm{[MASK]},...,x_M)$.
		MLM aims to model the conditional likelihood $P(x_m|\widetilde{\bm x})$, and the loss function is:
		\begin{align}
			L_{MLM}=\log P(x_m|\widetilde{\bm x})
		\end{align}	
		
		The major difference from BERT is that our KMLM prioritizes knowledge tokens, which contain knowledge regarding product attributes and USPs, when selecting positions to mask and, in the case that the knowledge tokens make up less than 15\% of all tokens, randomly picks non-knowledge tokens to complete the masking. 
		
		\textbf{Knowledge-aware Masked Sequence-to-Sequence (KMS2S)}
		
		K-PLUG inherits the strong ability of language generation from the masked sequence-to-sequence (MS2S) objective.
		The encoder takes a sentence with a masked fragment (several consecutive tokens) as the input, and the decoder predicts this masked fragment conditioned on the encoder representations (\textit{e.g.}, ``The company [MASK]  [MASK] [MASK]  the foot of a hill.'').
		
		Given a text sequence ${\bm x} = (x_1,...,x_u,...,x_v,...,x_M)$, a masked sequence $\widetilde{\bm x} = (x_1,...,\rm{[MASK]},...,\rm{[MASK]},...,x_M)$ is produced by replacing the span $\bm x_{u:v}$, ranging from $x_u$ to $x_v$, with the [MASK] token.  MS2S aims to model  $P(\bm x_{u:v}|\widetilde{\bm x})$, which can be further factorized into a product $P(\bm x_{u:v}|\widetilde{\bm x}) = \prod_{t=u}^vP(x_t|\widetilde{\bm x})$ according to the chain rule. The loss function is:
		\begin{align}
			L_{MS2S}=\sum_{t=u}^v\log P(x_t|\widetilde{\bm x})
		\end{align}	
		
		We set the  length of the masked span as  30\% of the length of the original text sequence. Similar to KMLM, KMS2S prioritizes the masking of text spans that cover knowledge tokens.

		\textbf{Product Entity Aspect Boundary Detection (PEABD)} 
		
		A product description usually contains multiple  product entity aspects. 
		Existing work~\citep{li-aaai2020-product} proves that product  aspects influence the quality of product summaries from the views of importance, non-redundancy, and readability, which are not directly taken into account in language modeling.
		In order to train a model that understands product aspects, we leverage the PEABD objective to detect boundaries between the product entity aspects. 
		It is essentially a sequence labeling task based on the representations of K-PLUG's top encoder layer. 
		
		Given a text sequence ${\bm x} = (x_1,...,x_M)$, the encoder of K-PLUG outputs a sequence ${\bm h} = (h_1,...,h_M)$, which is fed into a softmax layer, and  generates a probability sequence ${\bm y}$. The loss function is:
		\begin{align}
			L_{PEABD}=-\sum\limits_t\hat{y_t}\log y_t
		\end{align}	
		where ${\bm y}\in\{[0,1]\}$ are the ground-truth labels for the aspect boundary detection task.
		
		\textbf{Product Entity Category Classification (PECC)} 
		
		Product entity categories are the backbones for constructing taxonomies~\citep{luo2020alicoco,dong2020autoknow}.
		Each product description document corresponds to one of the 40 categories included in our corpus, such as \textit{Clothing}, \textit{Bags}, \textit{Home Appliances}, \textit{Shoes}, \textit{Foods},  etc. 
		Identifying the product entity categories accurately is the prerequisite for creating an output that is consistent with the input.
		
		Given a text sequence ${\bm x} = (x_1,...,x_M)$, a softmax layer outputs the classification score, $y$,  based on the representation of the encoder classification token, [CLS]. The loss function maximizes the model's probability of outputting the true product entity category as follows:
		\begin{align}
			L_{PECC}=-\hat{y}\log y
		\end{align}	
		where $\hat{y}$ is the ground-truth product category.
			
		\textbf{Product Entity Aspect Summary Generation (PEASG)} 
		
		Inspired by PEGASUS~\citep{zhang2019pegasus}, which proves that using a pre-training objective that more closely resembles the downstream task leads to better and faster fine-tuning performance, we propose a PEASG objective to generate a summary from the description of a product entity aspect.
		Unlike extracted gap-sentences generation in PEGASUS, our method constructs a more realistic summary generation task  because the aspect summary naturally exists in our pre-training data.

		Given an aspect description sequence ${\bm x} = (x_1,...,x_M)$, and an aspect summary sequence ${\bm y} = (y_1,...,y_T)$, PEASG aims to model the conditional likelihood $P(\bm y|\bm x)$. The loss function is:
		\begin{align}
			L_{PEASG}=\sum_t\log P(y_t|\bm x,\bm y_{<t})
		\end{align}

		Overall, the pre-training loss is the sum of  above-mentioned loss functions:
		\begin{align}
			L_{K-PLUG}=&L_{MLM}+L_{MS2S}+L_{PEABD}\nonumber\\
			&+L_{PECC}+L_{PEASG}
		\end{align}

		\section{Experiments and Results}
		
		\subsection{Pre-trained Model Variants}
		
		To evaluate  the effectiveness of pre-training with domain-specific data and with  domain-specific knowledge separately, we implement pre-training experiments with two model variants: C-PLUG and E-PLUG, whose configurations are the same as that of K-PLUG. 
		\begin{itemize}
			\item \textbf{C-PLUG}  is a pre-trained language model with the original objectives of MLM and MS2S, trained with a general pre-training corpus,  CLUE~\citep{xu2020clue}, which contains 30GB of raw text with around 8B Chinese words.
			\item \textbf{E-PLUG}  is a pre-trained language model with the original objectives of MLM and MS2S, trained with our collected e-commerce domain-specific corpus.
		\end{itemize}
		
		\subsection{Downstream Tasks}
		We fine-tune K-PLUG on three  downstream tasks: e-commerce KB completion, abstractive product  summarization, and multi-turn  dialogue. 
		The e-commerce KB completion task involves the prediction of product attributes and values given product information.
		The abstractive product summarization task requires the model to generate a product summary from textual product  description.
		The multi-turn  dialogue task aims to output the response by utilizing a multi-turn dialogue context. 
		The domain-specific knowledge we defined in this paper is essential for these tasks.
		
		\subsubsection{E-commerce KB Completion}
		\textbf{Task Definition.}
		E-commerce KB provides abundant product information that is in the form of  (\textit{product entity}, \textit{product attribute}, \textit{attribute value}), such as (\textit{pid\#133443}, \textit{Material}, \textit{Copper Aluminum}). For the E-commerce KB completion task, the input is a textual product description for a given product, and the output is the product attribute values.

		\textbf{Dataset.}
		We conduct experiments on the dataset of MEPAVE~\citep{zhu2020multimodal}. This dataset is collected from a major Chinese e-commerce platform, which consists of 87,194 instances annotated with the position of attribute values mentioned in the product descriptions. There are totally 26 types of product attributes such as \textit{Material}, \textit{Collar Type}, \textit{Color}, etc. 
		The training, validation, and testing sets contain 71,194/8,000/8,000 instances, respectively.
		
		\textbf{Model.}
		We consider the e-commerce KB completion task as a sequence labeling task that tags the input word sequence $x = (x_1, ..., x_N )$ with the label sequence $y = (y_1, ..., y_N )$ in the BIO format. For example, for the input sentence ``\textit{A bright yellow collar}'', the corresponding labels for \textit{``bright''}
		and ``\textit{yellow}'' are \textit{Color-B} and \textit{Color-I}, respectively, and \textit{O} for the other tokens.
		For an input sequence, K-PLUG outputs an encoding representation sequence, and a linear classification layer with the softmax predicts the label for each input token based on the encoding representation.

		\textbf{Baselines.} 
		\begin{itemize}
			\item \textbf{ScalingUp}~\citep{XuWMJL19} adopts BiLSTM, CRF, and attention mechanism to extract attributes.
			\item \textbf{JAVE}~\citep{zhu2020multimodal} is a joint attribute and value extraction model based on a pre-trained BERT.
			\item \textbf{M-JAVE}~\citep{zhu2020multimodal} is a multimodal JAVE model, which additionally utilizes product image information.
		\end{itemize}

\begin{table}\small	
	\begin{center}
		\begin{tabular}{lccc}
			\toprule
			\textbf{Model} & \textbf{P} & \textbf{R}& \textbf{F1}\\
			\midrule
			LSTM &79.68&86.43& 82.92 \\
			ScalingUp&65.48&93.78& 77.12 \\
			BERT &78.27&88.62& 83.12 \\
			JAVE &80.27&89.82& 84.78 \\
			M-JAVE &83.49&90.94& 87.17 \\
			\midrule
			C-PLUG &89.79&96.47& 93.02 \\
			E-PLUG  &89.91&96.75& 93.20 \\
			K-PLUG &\textbf{93.58}&\textbf{97.92}& \textbf{95.97} \\
			\bottomrule
		\end{tabular}
	\end{center}
	\caption{Experimental results with the F1 score for the e-commerce KB completion task. The results in the first block are taken from \citet{zhu2020multimodal}.}\label{tab:res_ekbc}
\end{table}

\textbf{Result.} 
Table~\ref{tab:res_ekbc} shows the experimental results. We observe that our K-PLUG performs better than baselines.
C-PLUG achieves significantly better performance than BERT, which indicates that MS2S can also benefit the NLU task.
E-PLUG outperforms C-PLUG, showing that training with domain-specific corpus is helpful.
K-PLUG further exhibits a 2.51\% improvement compared with E-PLUG.
In short, we can conclude that the improvement is due to both the domain-specific pre-training data and knowledge-injected pre-training objectives.

\subsubsection{Abstractive Product  Summarization}

		\textbf{Task Definition.}
		Abstractive product  summarization task aims to capture the most attractive information of a product that resonates with potential purchasers. The input for this task is a product description, and the output is a condensed product summary.
		
		\textbf{Dataset.}
		We perform experiments on the CEPSUM dataset~\citep{li-aaai2020-product}, which contains 1.4 million instances collected from a major Chinese e-commerce platform, covering three categories of product: \textit{Home Appliances}, \textit{Clothing}, and \textit{Cases \& Bags}. Each instance in the dataset is a (product information, product summary) pair, and the product information contains an image, a title, and other product descriptions. In our work, we do not consider the visual information of products.
		Notice that the task of abstractive product summarization and product entity aspect summary generation   (PEASG) are partly different.
		The abstractive product summarization task aims to generate a complete and cohesive product summary given a detailed product description. Given a product aspect description, PEASG aims to produce an aspect summary that basically consists of condensed USPs. In addition, for abstractive product summarization task, the average length of the product summaries is 79, while the lengths of the product aspect summaries are less than 10 in general.

		\textbf{Model.}
		Abstractive product  summarization task is an NLG task that takes the product description as the input and product summary as the output. 
		
		\textbf{Baselines.} 
		\begin{itemize}
			\item \textbf{LexRank}~\citep{abs-1109-2128} is a graph-based extractive summarization method.
			\item \textbf{Seq2seq}~\citep{bahdanau2014neural} is a standard seq2seq model with an attention mechanism.	
			\item \textbf{Pointer-Generator (PG)}~\citep{see2017get} is a seq2seq model with a copying mechanism.
			\item \textbf{Aspect MMPG}~\citep{li-aaai2020-product} is the-state-of-the-art method for  abstractive product summarization, taking both textual and visual product information as the input.
		\end{itemize}
		
		\textbf{Result.}
		Table~\ref{tab:res_esum} shows the experimental results, including ROUGE-1 (RG-1), ROUGE-2 (RG-2), and ROUGE-L (RG-L) F1 scores~\citep{Lin2003Automatic}.
		K-PLUG clearly performs better than other text-based methods. E-commerce knowledge plays a significant role in the abstractive product  summarization task, and domain-specific pre-training data and knowledge-injected pre-training objectives both enhance the model.
		K-PLUG achieves comparable results with the multimodal model, Aspect MMPG.
		The work of \citet{li-aaai2020-product} suggests that product images are essential for this task, and we will advance K-PLUG with multimodal information in the future.
\begin{table*}\small	
	\begin{center}
		\begin{tabular}{l|lll|lll|lll}
			\toprule
			\multirow{2}[0]{*}{\textbf{Model}} &\multicolumn{3}{c|}{\textbf{Home Applications}} & \multicolumn{3}{c|}{\textbf{Clothing}}&\multicolumn{3}{c}{\textbf{Cases\&Bags}} \\
			& \multicolumn{1}{l}{\textbf{RG-1}} & \multicolumn{1}{l}{\textbf{RG-2}} & \multicolumn{1}{l|}{\textbf{RG-L}}  & \multicolumn{1}{l}{\textbf{RG-1}} & \multicolumn{1}{l}{\textbf{RG-2}} & \multicolumn{1}{l|}{\textbf{RG-L}}& \multicolumn{1}{l}{\textbf{RG-1}} & \multicolumn{1}{l}{\textbf{RG-2}} & \multicolumn{1}{l}{\textbf{RG-L}} \\
			\midrule
			LexRank & 24.06 & 10.01 & 18.19 & 26.87 & 9.01  & 17.76 & 27.09 & 9.87  & 18.03 \\
			Seq2seq & 21.57 & 7.18  & 17.61 & 23.05 & 6.84  & 16.82 & 23.18 & 6.94  & 17.29 \\
			MASS  & 28.19 & 8.02  & 18.73 & 26.73 & 8.03  & 17.72 & 27.19 & 9.03  & 18.17 \\
			PG & 31.11 &10.93  &21.11& 29.11 & 9.24  &19.92& 31.31  & 10.27&21.79 \\
			\midrule
			Aspect MMPG*& 34.36& 12.52 &22.35& 31.93 & 11.09 &21.54& 33.78 & 12.51&22.43\\
			\midrule
			C-PLUG & 32.75 & 11.62 &21.76& 31.73 & 10.86  &20.37& 32.04 & 10.75&21.85 \\
			E-PLUG  & 33.11 & 12.07 &22.01& 32.61 & 11.03  &20.98& 32.37 & 11.14& 21.98\\
			K-PLUG & \textbf{33.56} & \textbf{12.50} &\textbf{22.15}& \textbf{33.00} & \textbf{11.24}   &\textbf{21.43}& \textbf{33.87} & \textbf{11.83} &\textbf{22.35}\\
			\bottomrule
		\end{tabular}
	\end{center}
	\caption{Experimental results with the ROUGE score for the abstractive product summarization task.
		The results in bold are the best performances among the models taking only texts as the input, and * denotes the model taking both product images and texts as the input. The results in the first and second blocks are taken from \citet{li-aaai2020-product}.}\label{tab:res_esum}
\end{table*}

\textbf{Human Evaluation.}
To help understand whether the knowledge has been learned during the pre-training, we conduct a knowledge-oriented human evaluation on 100 samples from the test set of CEPSUM dataset.
Three experienced annotators are involved to determine whether K-PLUG outperforms E-PLUG with respect to 
(1) KB: whether the model provides details about product attributes,
(2) Aspect: whether the model mentions distinctive product aspects,
(3) Category:  whether the model describes the correct product category,
and (4) USPs: whether the model generate proper USPs.
The results are shown in Table~\ref{tab:h}.
We can conclude that K-PLUG can learn the knowledge better than E-PLUG (p-value $<$ 0.01 for t-test). Kappa values~\citep{fleiss1971measuring} confirm the consistency for different annotators.
		
\begin{table}\small		
	\begin{center}
		\begin{tabular}{cccc}
			\toprule
			\multicolumn{2}{c}{KB}  & \multicolumn{2}{c}{Aspect}\\
			\midrule
			Win/Lose/Tie &Kappa &Win/Lose/Tie & Kappa\\	
			\midrule
			32.67/11.00/56.33 & 0.515& 32.67/12.00/55.33 &0.441\\		
			\midrule
			\midrule
			\multicolumn{2}{c}{Category}  & \multicolumn{2}{c}{USPs}\\
			\midrule
			Win/Lose/Tie &Kappa &Win/Lose/Tie & Kappa\\	
			\midrule
		    25.33/7.00/67.67 & 0.612& 28.67/9.33/62.00 &0.428\\		
			\bottomrule
		\end{tabular}%
	\end{center}
	\caption{Human evaluation results (\%). ``Win'' denotes that the generated summary of K-PLUG is better than E-PLUG. }
	\label{tab:h}%
\end{table}%

		\subsubsection{Multi-Turn Dialogue}  
		\textbf{Task Definition.}
		The multi-turn dialogue task aims to output a response based on the multi-turn dialogue context~\citep{shum2018eliza}. The input for this task is the dialogue context consisting of previous question answering, and the output is the response  to the last question.
		
		\textbf{Dataset.}
		We conduct experiments on two datasets of \textbf{JDDC}~\citep{chen2019jddc} and \textbf{ECD}~\citep{zhang2018modeling}.
		\textbf{JDDC} is collected from the conversations between users and customer service staffs from a popular e-commerce website in China and  contains 289 different intents, which are the goals of a dialogue, such as updating addresses, inquiring prices, etc, from after-sales assistance. There are 1,024,196 multi-turn sessions and 20,451,337 utterances in total. The average number of turns for each session is 20, and the average tokens per utterance is about 7.4. After pre-processing, the training, validation, and testing sets include 1,522,859/5,000/5,000 (dialogue context, response) pairs, respectively.
		\textbf{ECD} is collected from another popular e-commerce website in China and covers over 5 types of conversations based on 20 commodities. Additionally, for each ground-truth response, negative responses are provided for discriminative learning.
		The training, validation, and testing sets include 1,000,000/10,000/10,000 (dialogue context, response) pairs, respectively.

		\textbf{Model.}
		We test with two types of K-PLUG: retrieval-based K-PLUG on the ECD dataset and generative-based K-PLUG on the JDDC dataset.
		For the retrieval-based approach, we concatenate the dialogue context and use [SEP] token to separate context and response. The [CLS] representation is fed into the output layer for classification. The generative-based approach is a sequence-to-sequence model, which is the same as the model adopted in the abstractive product  summarization task.

	\textbf{Baselines.}
		The baselines also include both the retrieval-based (BM25, CNN, BiLSTM, and BERT) and generative-based approaches.
		Other baselines are as follows.
		\begin{itemize}
			\item \textbf{SMN}~\citep{wu2016sequential}  matches a response with each utterance in the context.
			\item \textbf{DUA}~\citep{zhang2018modeling} is a deep utterance aggregation model based on the fine-grained context representations. 
			\item \textbf{DAM}~\citep{zhou2018multi} matches a response with the context based using dependency information based on self-attention and cross-attention.
			\item \textbf{IoI}~\citep{tao2019one} is a deep matching model by stacking multiple interactions blocks between utterance and response.
			\item \textbf{MSN}~\citep{yuan2019multi} selects relevant context and generates better context representations with the selected context.
		\end{itemize}

\textbf{Result.}
Table~\ref{tab:res_jddc} and \ref{tab:res_ecd} show the experimental results on the JDDC  and ECD datasets, respectively. 
We report ROUGE-L (RG-L) F1, BLEU, and recall at position $k$ in $n$ candidates ($R_n@k$).
We can observe that, both on the retrieval-based and generative-based tasks,  K-PLUG achieves new state-of-the-art results, and e-commerce knowledge presents consistent improvements. 
K-PLUG is evidently superior to BERT, possibly due to BERT's lack of domain-specific knowledge for pre-training with the general MLM objective.
\begin{table}\small
	\begin{center}
		\begin{tabular}{l|cc}
			\toprule
			\textbf{Model}& \multicolumn{1}{l}{\textbf{RG-L}} & \multicolumn{1}{l}{\textbf{BLEU}} \\
			\midrule
			BM25 & 19.47 & 9.94 \\
			BERT & 19.90  & 10.27 \\
			Seq2Seq & 22.17 & 14.15 \\
			PG & 23.62 & 14.27 \\
			\midrule
			C-PLUG & 25.47 & 16.75 \\
			E-PLUG  & 25.93 & 17.12 \\
			K-PLUG & \textbf{26.60}  & \textbf{17.80} \\
			\bottomrule
		\end{tabular}
	\end{center}
	\caption{Experimental results  for the multi-turn conversation task on the JDDC dataset. The results in the first block are taken from \citet{chen2019jddc}.}\label{tab:res_jddc}
\end{table}

\begin{table}\small
	\begin{center}
	\begin{tabular}{l|ccc}
	\toprule
	\textbf{Model} & \multicolumn{1}{c}{\bm{$R_{10}@1$}} & \multicolumn{1}{c}{\bm{$R_{10}@2$}} & \multicolumn{1}{c}{\bm{{$R_{10}@5$}}} \\
	\midrule
	CNN  & 32.8 & 51.5 & 79.2 \\
	BiLSTM & 35.5 & 52.5 & 82.5 \\
	SMN  & 45.3 & 65.4 & 88.6 \\
	DUA  & 50.1 & 70.0 & 92.1 \\
	DAM  & 52.6	& 72.7 & 93.3 \\
	IoI-local&56.3	&76.8	&95.0\\
	MSN  &60.6	&77.0	&93.7\\
	BERT  & 54.3 & 73.4 & 94.3 \\
	\midrule
	C-PLUG & 62.7 & 76.8 & 95.0 \\
	E-PLUG  & 65.8 & 80.1 & \ 95.6  \\
	K-PLUG & \textbf{73.5} & \textbf{82.9} & \textbf{96.4} \\
	\bottomrule
\end{tabular}%
	\end{center}
\caption{Experimental results  for the multi-turn conversation task on the ECD dataset. The results in the first block are taken from \citet{zhang2018modeling}.}\label{tab:res_ecd}
\end{table}

\textbf{Human Evaluation.}
We further perform a human evaluation on the JDDC dataset.
We randomly choose 100 samples from the test set, and three  annotators are involved to determine whether K-PLUG outperforms E-PLUG with respect to (1) relevance between the response and the contexts and (2) readability of the response.
The results are shown in Table~\ref{tab:5}.
We can see that the percentage of ``Win'', which denotes that the results of K-PLUG is better than E-PLUG, is significantly larger than ``Lose'' (p-value $<$ 0.01 for t-test). 

\begin{table}\small		
	\begin{center}
			\begin{tabular}{cccc}
				\toprule
				\multicolumn{2}{c}{Relevance}  & \multicolumn{2}{c}{Readability}\\
				\midrule
				Win/Lose/Tie &Kappa &Win/Lose/Tie & Kappa\\	
				\midrule
				29.00/21.00/50.00 & 0.428& 7.00/2.00/91.00 &0.479\\		
				\bottomrule
			\end{tabular}%
			\end{center}
			\caption{Human evaluation results (\%). ``Win'' denotes that the generated response of K-PLUG is better than E-PLUG. }
			\label{tab:5}%
\end{table}%

\subsection{Ablation Studies}
		
		To better understand our model, we perform ablation experiments to  study the effects of different pre-training objectives.
		
		\textbf{Result.} 
		The ablation results are shown in Table~\ref{tab:res_abla}.
		We can conclude that the lack of any pre-training objective hurts performance across all the tasks. KMS2S is the most effective objective for the abstractive product  summarization and generative conversation tasks since this objective is highly close to the essence of NLG. Product-aspect-related objectives, \textit{i.e.}, PEABD and PEASG, contribute much to the abstractive product  summarization task, which proves  that this task requires comprehensively understanding the product description from the view of product aspects, going beyond individual tokens.
\begin{table*}\small
	\centering	
	\begin{tabular}{l|c|cccccc|cc}
		\toprule
		\multirow{3}[0]{*}{\textbf{Model}} & \textbf{KB} & \multicolumn{6}{c|}{\textbf{Abstractive Product  Summarization}}     & \multicolumn{2}{c}{\textbf{Multi-Turn}} \\
		& \textbf{Completion} & \multicolumn{2}{c}{\textbf{Home Applications}} & \multicolumn{2}{c}{\textbf{Clothing}} & \multicolumn{2}{c|}{\textbf{Cases\&Bags}} & \multicolumn{2}{c}{\textbf{Conversation}} \\
		& \textbf{F1}    & \textbf{RG-1} & \textbf{RG-2} & \textbf{RG-1} & \textbf{RG-2}&  \textbf{RG-1} & \textbf{RG-2} & \textbf{RG-L}  & \textbf{BLEU} \\
		\midrule
		K-PLUG & \textbf{95.97} & \textbf{33.56} & \textbf{12.50}  & \textbf{33.00}    & \textbf{11.24} & \textbf{33.87} & \textbf{11.83} & \textbf{26.60}  & \textbf{17.80} \\
		-KMLM  & 95.88 & 33.52 & 12.43 & 32.87 & 11.20  & 33.75 & 11.70  & 26.43 & 17.62 \\
		-KMS2S & 95.76 & 33.13 & 12.14 & 32.12 & 10.97 & 33.74 & 11.43	 & 25.82 & 16.97 \\
		-PEABD  & 95.89 & 33.26  & 12.30  & 32.96 & 11.14 & 33.69	 & 11.17 & 26.07 & 17.58 \\
		-PECC   & 95.59 & 33.24 & 12.17 & 32.25 & 11.12 & 33.59 & 11.18 & 26.02 & 17.16 \\
		-PEASG  & 95.48 & 33.39 & 12.36 & 32.57 & 11.16 & 33.78 & 11.45 & 26.12 & 17.38	 \\
		\bottomrule
	\end{tabular}%
	\caption{Experimental results  for ablation studies.}
	\label{tab:res_abla}%
\end{table*}%
		
\section{Conclusion}
		
		We present a knowledge-injected pre-trained  model (K-PLUG) that is a powerful domain-specific language model trained on a large-scale e-commerce corpus designed to capture e-commerce knowledge, including e-commerce KB, product aspects, product categories, and USPs.
		The pre-training framework combines masked language model and masked seq2seq with novel objectives formulated as product aspect boundary detection, product aspect summary generation, and product category classification  tasks.
		Our proposed model demonstrates strong performances on  both natural language understanding and generation downstream tasks, including  e-commerce KB completion, abstractive product  summarization, and multi-turn dialogue.
		
		\normalem
		\bibliography{custom}

\begin{thebibliography}{44}
\expandafter\ifx\csname natexlab\endcsname\relax\def\natexlab#1{#1}\fi

\bibitem[{Bahdanau et~al.(2015)Bahdanau, Cho, and Bengio}]{bahdanau2014neural}
Dzmitry Bahdanau, Kyunghyun Cho, and Yoshua Bengio. 2015.
\newblock \href {https://arxiv.org/abs/1409.0473} {Neural machine translation
  by jointly learning to align and translate}.
\newblock In \emph{3rd International Conference on Learning Representations,
  (ICLR)}.

\bibitem[{Chen et~al.(2020)Chen, Liu, Shen, Yuan, Zhou, Wu, He, and
  Zhou}]{chen2019jddc}
Meng Chen, Ruixue Liu, Lei Shen, Shaozu Yuan, Jingyan Zhou, Youzheng Wu,
  Xiaodong He, and Bowen Zhou. 2020.
\newblock \href {https://www.aclweb.org/anthology/2020.lrec-1.58} {The {JDDC}
  corpus: A large-scale multi-turn {C}hinese dialogue dataset for {E}-commerce
  customer service}.
\newblock In \emph{Proceedings of the 12th Language Resources and Evaluation
  Conference}, pages 459--466.

\bibitem[{Devlin et~al.(2019)Devlin, Chang, Lee, and
  Toutanova}]{devlin2018bert}
Jacob Devlin, Ming-Wei Chang, Kenton Lee, and Kristina Toutanova. 2019.
\newblock \href {https://www.aclweb.org/anthology/N19-1423/} {{BERT}:
  Pre-training of deep bidirectional transformers for language understanding}.
\newblock In \emph{Proceedings of the 2019 Conference of the North {A}merican
  Chapter of the Association for Computational Linguistics: Human Language
  Technologies (NAACL)}, pages 4171--4186.

\bibitem[{Dong et~al.(2019)Dong, Yang, Wang, Wei, Liu, Wang, Gao, Zhou, and
  Hon}]{UNILM}
Li~Dong, Nan Yang, Wenhui Wang, Furu Wei, Xiaodong Liu, Yu~Wang, Jianfeng Gao,
  Ming Zhou, and Hsiao{-}Wuen Hon. 2019.
\newblock \href
  {https://openreview.net/pdf/231668712e1f09679beadf07a5b7215bf64a8bd3.pdf}
  {Unified language model pre-training for natural language understanding and
  generation}.
\newblock In \emph{Advances in Neural Information Processing Systems 32: Annual
  Conference on Neural Information Processing Systems, (NeurIPS)}, pages
  13042--13054.

\bibitem[{Dong et~al.(2020)Dong, He, Kan, Li, Liang, Ma, Xu, Zhang, Zhao,
  Blanco~Saldana et~al.}]{dong2020autoknow}
Xin~Luna Dong, Xiang He, Andrey Kan, Xian Li, Yan Liang, Jun Ma, Yifan~Ethan
  Xu, Chenwei Zhang, Tong Zhao, Gabriel Blanco~Saldana, et~al. 2020.
\newblock \href {https://dl.acm.org/doi/abs/10.1145/3394486.3403323}
  {Auto{K}now: Self-driving knowledge collection for products of thousands of
  types}.
\newblock In \emph{Proceedings of the 26th ACM SIGKDD International Conference
  on Knowledge Discovery \& Data Mining (KDD)}, pages 2724--2734.

\bibitem[{Erkan and Radev(2004)}]{abs-1109-2128}
G{\"u}nes Erkan and Dragomir~R Radev. 2004.
\newblock \href {https://www.jair.org/index.php/jair/article/view/10396}
  {Lex{R}ank: Graph-based lexical centrality as salience in text
  summarization}.
\newblock \emph{Journal of Artificial Intelligence Research (JAIR)},
  22:457--479.

\bibitem[{Fleiss(1971)}]{fleiss1971measuring}
Joseph~L Fleiss. 1971.
\newblock \href {https://psycnet.apa.org/record/1972-05083-001} {Measuring
  nominal scale agreement among many raters}.
\newblock \emph{Psychological bulletin}, 76(5):378.

\bibitem[{Hendrycks and Gimpel(2016)}]{hendrycks2016gaussian}
Dan Hendrycks and Kevin Gimpel. 2016.
\newblock \href {https://arxiv.org/abs/1606.08415} {Gaussian error linear units
  (gelus)}.
\newblock \emph{arXiv preprint arXiv:1606.08415}.

\bibitem[{Ke et~al.(2020)Ke, Ji, Liu, Zhu, and Huang}]{ke2019sentilr}
Pei Ke, Haozhe Ji, Siyang Liu, Xiaoyan Zhu, and Minlie Huang. 2020.
\newblock \href {https://www.aclweb.org/anthology/2020.emnlp-main.567}
  {{S}enti{LARE}: Sentiment-aware language representation learning with
  linguistic knowledge}.
\newblock In \emph{Proceedings of the 2020 Conference on Empirical Methods in
  Natural Language Processing (EMNLP)}, pages 6975--6988.

\bibitem[{Kingma and Ba(2015)}]{kingma2014adam}
Diederik~P Kingma and Jimmy Ba. 2015.
\newblock \href {https://openreview.net/forum?id=8gmWwjFyLj} {Adam: A method
  for stochastic optimization}.
\newblock In \emph{3rd International Conference on Learning Representations,
  (ICLR)}.

\bibitem[{Lauscher et~al.(2019)Lauscher, Vuli{\'c}, Ponti, Korhonen, and
  Glava{\v{s}}}]{lauscher2019informing}
Anne Lauscher, Ivan Vuli{\'c}, Edoardo~Maria Ponti, Anna Korhonen, and Goran
  Glava{\v{s}}. 2019.
\newblock \href {https://arxiv.org/abs/1909.02339} {Specializing unsupervised
  pretraining models for word-level semantic similarity}.
\newblock \emph{arXiv preprint arXiv:1909.02339}.

\bibitem[{Levine et~al.(2020)Levine, Lenz, Dagan, Ram, Padnos, Sharir,
  Shalev-Shwartz, Shashua, and Shoham}]{levine2019sensebert}
Yoav Levine, Barak Lenz, Or~Dagan, Ori Ram, Dan Padnos, Or~Sharir, Shai
  Shalev-Shwartz, Amnon Shashua, and Yoav Shoham. 2020.
\newblock \href {https://www.aclweb.org/anthology/2020.acl-main.423}
  {{S}ense{BERT}: Driving some sense into {BERT}}.
\newblock In \emph{Proceedings of the 58th Annual Meeting of the Association
  for Computational Linguistics (ACL)}, pages 4656--4667.

\bibitem[{Lewis et~al.(2020)Lewis, Liu, Goyal, Ghazvininejad, Mohamed, Levy,
  Stoyanov, and Zettlemoyer}]{BART}
Mike Lewis, Yinhan Liu, Naman Goyal, Marjan Ghazvininejad, Abdelrahman Mohamed,
  Omer Levy, Veselin Stoyanov, and Luke Zettlemoyer. 2020.
\newblock \href {https://www.aclweb.org/anthology/2020.acl-main.703/} {{BART}:
  Denoising sequence-to-sequence pre-training for natural language generation,
  translation, and comprehension}.
\newblock In \emph{Proceedings of the 58th Annual Meeting of the Association
  for Computational Linguistics (ACL)}, pages 7871--7880.

\bibitem[{Li et~al.(2020)Li, Yuan, Xu, Wu, He, and Zhou}]{li-aaai2020-product}
Haoran Li, Peng Yuan, Song Xu, Youzheng Wu, Xiaodong He, and Bowen Zhou. 2020.
\newblock \href {https://ojs.aaai.org/index.php/AAAI/article/view/6332}
  {Aspect-aware multimodal summarization for chinese e-commerce products}.
\newblock In \emph{Proceedings of the Thirty-Forth {AAAI} Conference on
  Artificial Intelligence (AAAI)}, pages 8188--8195.

\bibitem[{Lin and Hovy(2003)}]{Lin2003Automatic}
Chin-Yew Lin and Eduard Hovy. 2003.
\newblock \href {https://www.aclweb.org/anthology/N03-1020/} {Automatic
  evaluation of summaries using n-gram co-occurrence statistics}.
\newblock In \emph{Proceedings of the 2003 Human Language Technology Conference
  of the North {A}merican Chapter of the Association for Computational
  Linguistics (NAACL)}, pages 150--157.

\bibitem[{Liu et~al.(2019)Liu, Ott, Goyal, Du, Joshi, Chen, Levy, Lewis,
  Zettlemoyer, and Stoyanov}]{liu2019roberta}
Yinhan Liu, Myle Ott, Naman Goyal, Jingfei Du, Mandar Joshi, Danqi Chen, Omer
  Levy, Mike Lewis, Luke Zettlemoyer, and Veselin Stoyanov. 2019.
\newblock \href {https://arxiv.org/abs/1907.11692} {Ro{BERT}a: A robustly
  optimized {BERT} pretraining approach}.
\newblock \emph{arXiv preprint arXiv:1907.11692}.

\bibitem[{Logan~IV et~al.(2017)Logan~IV, Humeau, and
  Singh}]{logan2017multimodal}
Robert~L Logan~IV, Samuel Humeau, and Sameer Singh. 2017.
\newblock \href {https://arxiv.org/abs/1711.11118} {Multimodal attribute
  extraction}.
\newblock \emph{arXiv preprint arXiv:1711.11118}.

\bibitem[{Luo et~al.(2020)Luo, Liu, Yang, Bo, Cao, Wu, Li, Yang, and
  Zhu}]{luo2020alicoco}
Xusheng Luo, Luxin Liu, Yonghua Yang, Le~Bo, Yuanpeng Cao, Jinghang Wu, Qiang
  Li, Keping Yang, and Kenny~Q Zhu. 2020.
\newblock \href {https://dl.acm.org/doi/abs/10.1145/3318464.3386132}
  {Ali{C}o{C}o: Alibaba e-commerce cognitive concept net}.
\newblock In \emph{Proceedings of the 2020 ACM SIGMOD International Conference
  on Management of Data (SIGMOD)}, pages 313--327.

\bibitem[{Peters et~al.(2018)Peters, Neumann, Iyyer, Gardner, Clark, Lee, and
  Zettlemoyer}]{peters-etal-2018-deep}
Matthew Peters, Mark Neumann, Mohit Iyyer, Matt Gardner, Christopher Clark,
  Kenton Lee, and Luke Zettlemoyer. 2018.
\newblock \href {https://www.aclweb.org/anthology/N18-1202/} {Deep
  contextualized word representations}.
\newblock In \emph{Proceedings of the 2018 Conference of the North {A}merican
  Chapter of the Association for Computational Linguistics: Human Language
  Technologies (NAACL)}, pages 2227--2237.

\bibitem[{Peters et~al.(2019)Peters, Neumann, Logan, Schwartz, Joshi, Singh,
  and Smith}]{peters2019knowledge}
Matthew~E. Peters, Mark Neumann, Robert Logan, Roy Schwartz, Vidur Joshi,
  Sameer Singh, and Noah~A. Smith. 2019.
\newblock \href {https://www.aclweb.org/anthology/D19-1005/} {Knowledge
  enhanced contextual word representations}.
\newblock In \emph{Proceedings of the 2019 Conference on Empirical Methods in
  Natural Language Processing and the 9th International Joint Conference on
  Natural Language Processing (EMNLP-IJCNLP)}, pages 43--54.

\bibitem[{Qi et~al.(2020)Qi, Yan, Gong, Liu, Duan, Chen, Zhang, and
  Zhou}]{yan2020prophetnet}
Weizhen Qi, Yu~Yan, Yeyun Gong, Dayiheng Liu, Nan Duan, Jiusheng Chen, Ruofei
  Zhang, and Ming Zhou. 2020.
\newblock \href {https://www.aclweb.org/anthology/2020.findings-emnlp.217}
  {{P}rophet{N}et: Predicting future n-gram for
  sequence-to-{S}equence{P}re-training}.
\newblock In \emph{Findings of the Association for Computational Linguistics:
  EMNLP 2020}, pages 2401--2410.

\bibitem[{Radford et~al.(2018)Radford, Narasimhan, Salimans, and
  Sutskever}]{GPT}
Alec Radford, Karthik Narasimhan, Tim Salimans, and Ilya Sutskever. 2018.
\newblock \href
  {https://www.cs.ubc.ca/~amuham01/LING530/papers/radford2018improving.pdf}
  {Improving language understanding by generative pre-training}.

\bibitem[{Raffel et~al.(2020)Raffel, Shazeer, Roberts, Lee, Narang, Matena,
  Zhou, Li, and Liu}]{raffel2019exploring}
Colin Raffel, Noam Shazeer, Adam Roberts, Katherine Lee, Sharan Narang, Michael
  Matena, Yanqi Zhou, Wei Li, and Peter~J Liu. 2020.
\newblock \href {https://w.jmlr.org/papers/volume21/20-074/20-074.pdf}
  {Exploring the limits of transfer learning with a unified text-to-text
  transformer}.
\newblock \emph{Journal of Machine Learning Research}, 21:1--67.

\bibitem[{Reeves(1961)}]{reeves2017reality}
Rosser Reeves. 1961.
\newblock \href
  {https://books.google.com.sg/books?hl=zh-CN&lr=&id=P1AoDwAAQBAJ&oi=fnd&pg=PA3&dq=Reality+in+Advertising&ots=H2oPNKsGIt&sig=BVTtsdUbh6DLV1D3l2oRn4Ni2xo&redir_esc=y#v=onepage&q=Reality%20in%20Advertising&f=false}
  {\emph{Reality in advertising}}.
\newblock Knopf.

\bibitem[{See et~al.(2017)See, Liu, and Manning}]{see2017get}
Abigail See, Peter~J. Liu, and Christopher~D. Manning. 2017.
\newblock \href {https://www.aclweb.org/anthology/P17-1099/} {Get to the point:
  Summarization with pointer-generator networks}.
\newblock In \emph{Proceedings of the 55th Annual Meeting of the Association
  for Computational Linguistics (ACL)}, pages 1073--1083.

\bibitem[{Shum et~al.(2018)Shum, He, and Li}]{shum2018eliza}
Heung-Yeung Shum, Xiao-dong He, and Di~Li. 2018.
\newblock \href {https://link.springer.com/article/10.1631/FITEE.1700826} {From
  {E}liza to {X}iaoice: challenges and opportunities with social chatbots}.
\newblock \emph{Frontiers of Information Technology \& Electronic Engineering},
  19:10--26.

\bibitem[{Song et~al.(2019)Song, Tan, Qin, Lu, and Liu}]{song2019mass}
Kaitao Song, Xu~Tan, Tao Qin, Jianfeng Lu, and Tie{-}Yan Liu. 2019.
\newblock \href {https://openreview.net/forum?id=H1EkFhZd-H} {{MASS:} masked
  sequence to sequence pre-training for language generation}.
\newblock In \emph{Proceedings of the 36th International Conference on Machine
  Learning (ICML)}, pages 5926--5936.

\bibitem[{Sun et~al.(2019)Sun, Wang, Li, Feng, Chen, Zhang, Tian, Zhu, Tian,
  and Wu}]{sun2019ernie}
Yu~Sun, Shuohuan Wang, Yukun Li, Shikun Feng, Xuyi Chen, Han Zhang, Xin Tian,
  Danxiang Zhu, Hao Tian, and Hua Wu. 2019.
\newblock \href {https://arxiv.org/abs/1904.09223} {{ERNIE}: Enhanced
  representation through knowledge integration}.
\newblock \emph{arXiv preprint arXiv:1904.09223}.

\bibitem[{Tao et~al.(2019)Tao, Wu, Xu, Hu, Zhao, and Yan}]{tao2019one}
Chongyang Tao, Wei Wu, Can Xu, Wenpeng Hu, Dongyan Zhao, and Rui Yan. 2019.
\newblock \href {https://www.aclweb.org/anthology/P19-1001/} {One time of
  interaction may not be enough: Go deep with an interaction-over-interaction
  network for response selection in dialogues}.
\newblock In \emph{Proceedings of the 57th Annual Meeting of the Association
  for Computational Linguistics}, pages 1--11.

\bibitem[{Tian et~al.(2020)Tian, Gao, Xiao, Liu, He, Wu, Wang, and
  Wu}]{tian2020skep}
Hao Tian, Can Gao, Xinyan Xiao, Hao Liu, Bolei He, Hua Wu, Haifeng Wang, and
  Feng Wu. 2020.
\newblock \href {https://www.aclweb.org/anthology/2020.acl-main.374/} {{SKEP}:
  Sentiment knowledge enhanced pre-training for sentiment analysis}.
\newblock In \emph{Proceedings of the 58th Annual Meeting of the Association
  for Computational Linguistics (ACL)}, pages 4067--4076.

\bibitem[{Vaswani et~al.(2017)Vaswani, Shazeer, Parmar, Uszkoreit, Jones,
  Gomez, Kaiser, and Polosukhin}]{vaswani2017attention}
Ashish Vaswani, Noam Shazeer, Niki Parmar, Jakob Uszkoreit, Llion Jones,
  Aidan~N. Gomez, Lukasz Kaiser, and Illia Polosukhin. 2017.
\newblock \href {https://openreview.net/forum?id=H1ZGYPb_ZS} {Attention is all
  you need}.
\newblock In \emph{Advances in Neural Information Processing Systems 30: Annual
  Conference on Neural Information Processing Systems (NeurIPS)}, pages
  5998--6008.

\bibitem[{Wang et~al.(2020)Wang, Tang, Duan, Wei, Huang, Cao, Jiang, Zhou
  et~al.}]{wang2020k}
Ruize Wang, Duyu Tang, Nan Duan, Zhongyu Wei, Xuanjing Huang, Cuihong Cao,
  Daxin Jiang, Ming Zhou, et~al. 2020.
\newblock \href {https://arxiv.org/abs/2002.01808} {K-{A}dapter: Infusing
  knowledge into pre-trained models with adapters}.
\newblock \emph{arXiv preprint arXiv:2002.01808}.

\bibitem[{Wang et~al.(2019)Wang, Gao, Zhu, Liu, Li, and Tang}]{wang2019kepler}
Xiaozhi Wang, Tianyu Gao, Zhaocheng Zhu, Zhiyuan Liu, Juanzi Li, and Jian Tang.
  2019.
\newblock \href {https://arxiv.org/abs/1911.06136} {{KEPLER}: A unified model
  for knowledge embedding and pre-trained language representation}.
\newblock \emph{arXiv preprint arXiv:1911.06136}.

\bibitem[{Wu et~al.(2017)Wu, Wu, Xing, Zhou, and Li}]{wu2016sequential}
Yu~Wu, Wei Wu, Chen Xing, Ming Zhou, and Zhoujun Li. 2017.
\newblock \href {https://www.aclweb.org/anthology/P17-1046/} {Sequential
  matching network: A new architecture for multi-turn response selection in
  retrieval-based chatbots}.
\newblock In \emph{Proceedings of the 55th Annual Meeting of the Association
  for Computational Linguistics (ACL)}, pages 496--505.

\bibitem[{Xiong et~al.(2020)Xiong, Du, Wang, and
  Stoyanov}]{xiong2019pretrained}
Wenhan Xiong, Jingfei Du, William~Yang Wang, and Veselin Stoyanov. 2020.
\newblock \href {https://openreview.net/forum?id=BJlzm64tDH} {Pretrained
  encyclopedia: Weakly supervised knowledge-pretrained language model}.
\newblock In \emph{8th International Conference on Learning Representations
  (ICLR)}.

\bibitem[{Xu et~al.(2019)Xu, Wang, Mao, Jiang, and Lan}]{XuWMJL19}
Huimin Xu, Wenting Wang, Xin Mao, Xinyu Jiang, and Man Lan. 2019.
\newblock \href {https://www.aclweb.org/anthology/P19-1514/} {Scaling up open
  tagging from tens to thousands: Comprehension empowered attribute value
  extraction from product title}.
\newblock In \emph{Proceedings of the 57th Annual Meeting of the Association
  for Computational Linguistics (ACL)}, pages 5214--5223.

\bibitem[{Xu et~al.(2020)Xu, Hu, Zhang, Li, Cao, Li, Xu, Sun, Yu, Yu, Tian,
  Dong, Liu, Shi, Cui, Li, Zeng, Wang, Xie, Li, Patterson, Tian, Zhang, Zhou,
  Liu, Zhao, Zhao, Yue, Zhang, Yang, Richardson, and Lan}]{xu2020clue}
Liang Xu, Hai Hu, Xuanwei Zhang, Lu~Li, Chenjie Cao, Yudong Li, Yechen Xu, Kai
  Sun, Dian Yu, Cong Yu, Yin Tian, Qianqian Dong, Weitang Liu, Bo~Shi, Yiming
  Cui, Junyi Li, Jun Zeng, Rongzhao Wang, Weijian Xie, Yanting Li, Yina
  Patterson, Zuoyu Tian, Yiwen Zhang, He~Zhou, Shaoweihua Liu, Zhe Zhao, Qipeng
  Zhao, Cong Yue, Xinrui Zhang, Zhengliang Yang, Kyle Richardson, and Zhenzhong
  Lan. 2020.
\newblock \href {https://www.aclweb.org/anthology/2020.coling-main.419}
  {{CLUE}: A {C}hinese language understanding evaluation benchmark}.
\newblock In \emph{Proceedings of the 28th International Conference on
  Computational Linguistics}, pages 4762--4772.

\bibitem[{Yang et~al.(2019)Yang, Dai, Yang, Carbonell, Salakhutdinov, and
  Le}]{yang2019xlnet}
Zhilin Yang, Zihang Dai, Yiming Yang, Jaime Carbonell, Russ~R Salakhutdinov,
  and Quoc~V Le. 2019.
\newblock \href
  {https://papers.nips.cc/paper/2019/hash/dc6a7e655d7e5840e66733e9ee67cc69-Abstract.html}
  {{XLNET}: Generalized autoregressive pretraining for language understanding}.
\newblock In \emph{Advances in Neural Information Processing Systems 32: Annual
  Conference on Neural Information Processing Systems (NeurIPS)}, pages
  5753--5763.

\bibitem[{Yuan et~al.(2019)Yuan, Zhou, Li, Lv, Zhu, Han, and
  Hu}]{yuan2019multi}
Chunyuan Yuan, Wei Zhou, Mingming Li, Shangwen Lv, Fuqing Zhu, Jizhong Han, and
  Songlin Hu. 2019.
\newblock \href {https://www.aclweb.org/anthology/D19-1011/} {Multi-hop
  selector network for multi-turn response selection in retrieval-based
  chatbots}.
\newblock In \emph{Proceedings of the 2019 Conference on Empirical Methods in
  Natural Language Processing and the 9th International Joint Conference on
  Natural Language Processing (EMNLP-IJCNLP)}, pages 111--120.

\bibitem[{Zhang et~al.(2020)Zhang, Zhao, Saleh, and Liu}]{zhang2019pegasus}
Jingqing Zhang, Yao Zhao, Mohammad Saleh, and Peter~J Liu. 2020.
\newblock \href {http://proceedings.mlr.press/v119/zhang20ae.html} {{PEGASUS}:
  Pre-training with extracted gap-sentences for abstractive summarization}.
\newblock In \emph{Proceedings of the 37th International Conference on Machine
  Learning (ICML)}.

\bibitem[{Zhang et~al.(2019)Zhang, Han, Liu, Jiang, Sun, and
  Liu}]{zhang2019ernie}
Zhengyan Zhang, Xu~Han, Zhiyuan Liu, Xin Jiang, Maosong Sun, and Qun Liu. 2019.
\newblock \href {https://www.aclweb.org/anthology/P19-1139/} {{ERNIE}: Enhanced
  language representation with informative entities}.
\newblock In \emph{Proceedings of the 57th Annual Meeting of the Association
  for Computational Linguistics (ACL)}, pages 1441--1451.

\bibitem[{Zhang et~al.(2018)Zhang, Li, Zhu, Zhao, and Liu}]{zhang2018modeling}
Zhuosheng Zhang, Jiangtong Li, Pengfei Zhu, Hai Zhao, and Gongshen Liu. 2018.
\newblock \href {https://www.aclweb.org/anthology/C18-1317/} {Modeling
  multi-turn conversation with deep utterance aggregation}.
\newblock In \emph{Proceedings of the 27th International Conference on
  Computational Linguistics (COLING)}, pages 3740--3752.

\bibitem[{Zhou et~al.(2018)Zhou, Li, Dong, Liu, Chen, Zhao, Yu, and
  Wu}]{zhou2018multi}
Xiangyang Zhou, Lu~Li, Daxiang Dong, Yi~Liu, Ying Chen, Wayne~Xin Zhao, Dianhai
  Yu, and Hua Wu. 2018.
\newblock \href {https://www.aclweb.org/anthology/P18-1103/} {Multi-turn
  response selection for chatbots with deep attention matching network}.
\newblock In \emph{Proceedings of the 56th Annual Meeting of the Association
  for Computational Linguistics (Volume 1: Long Papers)}, pages 1118--1127.

\bibitem[{Zhu et~al.(2020)Zhu, Wang, Li, Wu, He, and Zhou}]{zhu2020multimodal}
Tiangang Zhu, Yue Wang, Haoran Li, Youzheng Wu, Xiaodong He, and Bowen Zhou.
  2020.
\newblock \href {https://www.aclweb.org/anthology/2020.emnlp-main.166/}
  {Multimodal joint attribute prediction and value extraction for e-commerce
  product}.
\newblock In \emph{Proceedings of the 2020 Conference on Empirical Methods in
  Natural Language Processing (EMNLP)}.

\end{thebibliography}
		\bibliographystyle{acl_natbib}

\appendix

\section{Appendix}

\subsection{Case studies}
We present some examples from the test set of each task, with comparisons of the ground-truth result and the outputs produced by the models of E-PLUG and K-PLUG.

\begin{table*}\small
	\centering
	\begin{tabular}{l|l}
		\toprule
		\multicolumn{1}{l|}{Ground-truth} & ECCO 防 滑 简 约 筒 [短 靴 ]$_{\text{靴筒高度}}$\\
		& (ECCO's non-slip simple [ankle boots]$_{\text{shaft height}}$) \\
		\midrule
		\multicolumn{1}{l|}{E-PLUG} & ECCO 防 滑 简 约 筒 短 靴$_{\text{鞋跟高度}}$ \\
		& (ECCO's non-slip simple [ankle boots]$_{\text{heel height}}$) \\
		\midrule
		\multicolumn{1}{l|}{K-PLUG} & ECCO 防 滑 简 约 筒 [短 靴 ]$_{\text{靴筒高度}}$\\
		& (ECCO's non-slip simple [ankle boots]$_{\text{shaft height}}$) \\
		\midrule
		\midrule
		\multicolumn{1}{l|}{Ground-truth} & a21 [运 动 风]$_{\text{风格}}$[撞 色]$_{\text{图案}}$ 风 衣 \\
		& (A21's [sports]$_{\text{style}}$ windbreaker jacket with [contrasting color]$_{\text{design}}$ style) \\
		\midrule
		\multicolumn{1}{l|}{E-PLUG} & a21 [运 动 风]$_{\text{风格}}$ [撞 色 风]$_{\text{风格}}$ 衣 \\
		& (A21's sports [windbreaker]$_{\text{style}}$ jacket with [contrasting color style]$_{\text{style}}$  \\
		\midrule
		\multicolumn{1}{l|}{K-PLUG} & a21 [运 动 风]$_{\text{风格}}$[撞 色]$_{\text{图案}}$ 风 衣 \\
		& (A21's sports [windbreaker]$_{\text{style}}$ jacket with [contrasting color]$_{\text{design}}$ style) \\
		\midrule
		\midrule
		\multicolumn{1}{l|}{Ground-truth} & 配 合 [微 弹]$_{\text{弹性}}$ 的 [棉 质]$_{\text{材质}}$ 面 料 手 感 柔 软 顺 滑 \\
		& (made from [low-strech]$_{\text{elasticity}}$ [cotton fabric]$_{\text{material}}$ for a silky smooth touch) \\
		\midrule
		\multicolumn{1}{l|}{E-PLUG} & 配 合 [微 弹]$_{\text{裤型}}$ 的 [棉 质]$_{\text{材质}}$ 面 料 手 感 柔 软 顺 滑 \\
		& (made from [low-strech]$_{\text{pants fit}}$ [cotton fabric]$_{\text{material}}$  for a silky smooth touch) \\
		\midrule
		\multicolumn{1}{l|}{K-PLUG} & 配 合 [微 弹]$_{\text{弹性}}$ 的 [棉 质]$_{\text{材质}}$ 面 料 手 感 柔 软 顺 滑 \\
		& (made from [low-strech]$_{\text{elasticity}}$ [cotton fabric]$_{\text{material}}$ for a silky smooth touch) \\
		\bottomrule
	\end{tabular}%
		\caption{Case study for the e-commerce KB complete task. The \textit{product attribute} and the corresponding \textit{attribute value} is presented as [\textit{attribute value}]$_{\text{\textit{product attribute}}}$. The K-PLUG model accurately complete the  e-commerce KB, while the E-PLUG model sometimes fails. The translation texts are given below the original examples.}
	\label{tab:sup_kbc}%
\end{table*}%

\begin{table*}\small
	\centering

	\begin{tabular}{l|l}
		\toprule
		\multicolumn{1}{l|}{Input } & 帅康大吸力智能触控欧式抽油烟机，不锈钢材质，品质大于所见，多重贴心，310pa大 \\
		\multicolumn{1}{l|}{Description} & 风压，17立方大吸力，不碰头设计，53db静音，动力强，拢烟面积大，只留美味，不 \\
		& 留烟味 … \\
		& (Sacon's European style kitchen range hood. Extra powerful ventilation.  Intelligent touch  \\
		& control. Stainless steel. Superior quality with extra reliability. 310 Pa air pressure. 17 $m^3$ \\
		& suction. Protection against head bumping. Limited noise of 53 dB.  Highly powered. Captures \\
		&  large volume of fumes. Leaves no cooking smells ...) \\
		\midrule
		\multicolumn{1}{l|}{Ground-truth} & 想要清新的厨房环境，那就试试这款\textcolor[RGB]{46,117,182}{\uwave{\textit{大吸力}}}油烟机吧！\textcolor[RGB]{84,130,52}{\underline{\textit{310pa}}}大风压与\textcolor[RGB]{84,130,52}{\underline{\textit{17立方米}}} 的大吸\\
		&力的完美搭配，能够瞬间吸附烹饪过程中产生的油烟，让您清新烹饪，不做“黄脸婆”！ \\
		& (This kitchen range hood with \textcolor[RGB]{46,117,182}{\uwave{\textit{extra suction and ventilation}}} is ideal for a clean and fresh  kitchen.\\
		&  Perfectly combining \textcolor[RGB]{84,130,52}{\underline{\textit{310pa}}} of air pressure and \textcolor[RGB]{84,130,52}{\underline{\textit{17$m^3$}}}  of suction power, it filters out grease before \\
		&   you realize it. Never have to worry again about keeping your loved ones from cooking fumes. )\\
		\midrule
		\multicolumn{1}{l|}{E-PLUG} & 这款帅康油烟机，采用优质的\textcolor[RGB]{84,130,52}{\underline{\textit{不锈钢}}}材质，精工细作，打造出简约大气的外观，尽显出 \\
		& 欧式的厨房魅力，具有强劲的\textcolor[RGB]{46,117,182}{\uwave{\textit{大吸力}}}，快速吸走烹饪产生的油烟，智能的触控面板设计，\\
		& 使操作更加方便。 \\
		& (This kitchen range hood from Sacon is made of  \textcolor[RGB]{84,130,52}{\underline{\textit{stainless steel}}} with cutting edge  technology. Its \\
		& design adds contemporary style to your European style kitchen. It  features \textcolor[RGB]{46,117,182}{\uwave{\textit{extra suction and}}} \\
		&  \textcolor[RGB]{46,117,182}{\uwave{\textit{ventilation}}} for immediate filtering of fumes and grease.  Its intelligent touch control panel makes\\
		&  it extremely easy to use. )\\
		\midrule
		\multicolumn{1}{l|}{K-PLUG} & 这款\textcolor[RGB]{84,130,52}{\underline{\textit{不锈钢}}}油烟机拥有\textcolor[RGB]{84,130,52}{\underline{\textit{17立方}}}\textcolor[RGB]{46,117,182}{\uwave{\textit{大吸力}}}，快速清除油烟，给您一个清新的厨房环境。\textcolor[RGB]{84,130,52}{\underline{\textit{310pa}}} \\
		& 大风压，保持排烟畅通，防止油烟倒灌。低至\textcolor[RGB]{84,130,52}{\underline{\textit{53分贝}}}的运行声音，让您告别厨房的嘈杂。 \\
		& (This  \textcolor[RGB]{84,130,52}{\underline{\textit{stainless steel}}} kitchen range hood features  \textcolor[RGB]{46,117,182}{\uwave{\textit{extra suction and ventilation}}} of  \textcolor[RGB]{84,130,52}{\underline{\textit{17 $m^3$}}} for \\
		&  immediate filtering of grease and fumes, leaving your kitchen fresh.  Its \textcolor[RGB]{84,130,52}{\underline{\textit{310 pa}}} air pressure \\
		&  ensures clear filtering without backflow. Its low working  noise of \textcolor[RGB]{84,130,52}{\underline{\textit{53 dB}}} keeps your kitchen nice\\
		&  and quiet.) \\
		\midrule
		\midrule
		\multicolumn{1}{l|}{Input } & 康宝8升四季控温型，经济适用，家用厨房卫生间淋浴洗澡强排式燃气热水器，厨房洗菜， \\
		\multicolumn{1}{l|}{Description} & 冬天洗漱再也不冻手，实打实的配置，高纯度无氧铜水箱，LED数字大屏，封闭式燃烧 \\
		& 系统，防患于未然，控温清晰可见，多重安全防护，无氧铜水箱，让每次沐浴都是全新 \\
		& 享受 ... \\
		& (Canbo's affordable 8L thermal control natural gas residential kitchen and bathroom water heater.  \\
		& Provides warm water for doing dishes and cleaning up in winter. High-quality accessories. High  \\
		& concentration oxygen-free copper tank. LED digital screen. Closed boiling system for safety. Easy  \\
		& temperature control. Multiple security checks.  Oxygen-free copper tank. Lets you enjoy your \\
		& showers ... )\\
		\midrule
		\multicolumn{1}{l|}{Ground-truth} & 采用高纯度\textcolor[RGB]{84,130,52}{\underline{\textit{无氧铜}}}水箱，增大了加热面积，高效速热且温度持久稳定。\textcolor[RGB]{46,117,182}{\uwave{\textit{封闭式燃烧技术}}}， \\
		& 提升了热效率，节能又高效。拥有控温功能，搭配\textcolor[RGB]{46,117,182}{\uwave{\textit{LED数字大屏}}}，让你能够根据需求来 \\
		& 随意调节水温。 \\
		& (High concentration \textcolor[RGB]{84,130,52}{\underline{\textit{oxygen-free}}}  tank with added heating area boils water quickly with  stability.  \\
		& \textcolor[RGB]{46,117,182}{\uwave{\textit{Closed boiling technique}}} increases efficiency and saves energy.  Thermal control and \textcolor[RGB]{46,117,182}{\uwave{\textit{LED digital}}} \\
		&  \textcolor[RGB]{46,117,182}{\uwave{\textit{ screen}}} for adjustable water temperature anytime. )\\
		\midrule
		\multicolumn{1}{l|}{E-PLUG} & 这款燃气热水器容量大，可以满足一年四季的用水需求。它采用\textcolor[RGB]{46,117,182}{\uwave{\textit{封闭式燃烧技术}}}，出热水的 \\
		& 速度很快，沐浴的时候不用着急的等待，而且它在低压状态下也能正常运行，老小区用户 \\
		& 也适用。 \\
		& (This gas water heater has a large volume and provides warm water year round.  It adopts the \textcolor[RGB]{46,117,182}{\uwave{\textit{closed  }}}\\
		&   \textcolor[RGB]{46,117,182}{\uwave{\textit{boiling technique}}} for faster warming up and shorter wait time in shower. It works in low pressure \\
		&   condition and is suitable for older neighborhoods. )\\
		\midrule
		\multicolumn{1}{l|}{K-PLUG} & 高配\textcolor[RGB]{46,117,182}{\uwave{\textit{LED数字大屏}}}，控温清晰可见，温度随时掌控于心。\textcolor[RGB]{46,117,182}{\uwave{\textit{封闭式燃烧技术}}}，快速度燃烧， \\
		& 加热效率更高。高纯度\textcolor[RGB]{84,130,52}{\underline{\textit{无氧铜}}}水箱，加热速度更快。 \\
		& (Equipped with \textcolor[RGB]{46,117,182}{\uwave{\textit{LED digital screen}}} for clear thermal control and temperature adjustment.  \textcolor[RGB]{46,117,182}{\uwave{\textit{Closed }}} \\
		& \textcolor[RGB]{46,117,182}{\uwave{\textit{boiling technique}}} speeds up warming and increases efficiency. High concentration  \textcolor[RGB]{84,130,52}{\underline{\textit{oxygen-free }}} \\
		&  \textcolor[RGB]{84,130,52}{\underline{\textit{ copper}}} tank for faster warming up.) \\
		\bottomrule
	\end{tabular}%
		\caption{Case study for the abstractive product summarization task (\textit{Home Applications} category). The K-PLUG model generates summaries describing more information about   \textcolor[RGB]{84,130,52}{\underline{\textit{e-commerce knowledge bases}}} and \textcolor[RGB]{46,117,182}{\uwave{\textit{unique selling propositions of product entities}}}. }
	\label{tab:sup_psum_home}%
\end{table*}%

\begin{table*}\small
	\centering

	\begin{tabular}{l|l}
		\toprule
		\multicolumn{1}{l|}{Input } & 劲霸男士t恤，夏季新品，撞色时尚，舒适，花卉印花，短袖，黑色，商务休闲，圆领， \\
		\multicolumn{1}{l|}{Description} & 夏季，时尚都市，短袖t恤，悬垂感和耐穿性好，时尚圆领，简约大气，多色可选，莫代尔 \\
		& 棉针织面料，柔滑触感，清凉舒爽，花卉印花结合，珠片绣花，时尚大方 ... \\
		& (K-Boxing's men's t-Shirt. This summer's new arrival. Contrasting colors. Soft and comfortable. \\
		& Flower print. Short-sleeve. Black. Business casual. Crewneck. Summer. Urban style.  Short-sleeve \\
		&  t-shirt. Naturally draping with high durability. Stylish crewneck. Basic and causal. Multiple \\
		&  color chioces. Modal knit fabric. Soft and elastic. Loose fit. Flower print.  Glitter embroidery. \\
		& Stylish and casual ....)  \\
		\midrule
		\multicolumn{1}{l|}{Reference} & 选用\textcolor[RGB]{84,130,52}{\underline{\textit{莫代尔棉}}}\textcolor[RGB]{84,130,52}{\underline{\textit{针织}}}面料，\textcolor[RGB]{46,117,182}{\uwave{\textit{手感柔滑}}}，上身\textcolor[RGB]{46,117,182}{\uwave{\textit{清凉舒爽}}}，悬垂感好，耐穿性强。胸前撞色花卉 \\
		& 印花结合\textcolor[RGB]{46,117,182}{\uwave{\textit{珠片绣花}}}，时尚又大方。经典\textcolor[RGB]{84,130,52}{\underline{\textit{圆领}}}设计，自然突显\textcolor[RGB]{46,117,182}{\uwave{\textit{型男风范}}}。 \\
		& (Made from \textcolor[RGB]{84,130,52}{\underline{\textit{modal knit fabric}}}. \textcolor[RGB]{46,117,182}{\uwave{\textit{Incredibly soft and elastic}}}. Features a \textcolor[RGB]{46,117,182}{\uwave{\textit{slightly looser fit}}}.  Natural \\
		&  draping with high durability. Decorated with flower print and  \textcolor[RGB]{46,117,182}{\uwave{\textit{glitter embroidery}}} for a stylish and \\
		&  casual look. Finished with a classy \textcolor[RGB]{84,130,52}{\underline{\textit{crewneck}}},  \textcolor[RGB]{46,117,182}{\uwave{\textit{appealing to the modern sense of fashion}}})\\
		\midrule
		\multicolumn{1}{l|}{E-PLUG} &  \textcolor[RGB]{84,130,52}{\underline{\textit{圆领}}}的领口更好的将颈部曲线所勾勒而出。水洗磨白的颜色平添了几分帅气与随性。让你 \\
		& 和夏季的风一起随行。让休闲舒适的田园感油然而生。 \\
		& (The \textcolor[RGB]{84,130,52}{\underline{\textit{crewneck}}} design helps in elongating the neck. The light wash color gives an  attractive yet \\
		& effortless look. It dresses you up in a country asethetic style, comfy and relaxed just like the summer\\
		&    breeze. )\\
		\midrule
		\multicolumn{1}{l|}{K-PLUG} & 采用\textcolor[RGB]{84,130,52}{\underline{\textit{莫代尔棉}}}\textcolor[RGB]{84,130,52}{\underline{\textit{针织}}}面料，\textcolor[RGB]{46,117,182}{\uwave{\textit{柔滑触感}}}，\textcolor[RGB]{46,117,182}{\uwave{\textit{清凉舒爽}}}，花卉印花结合\textcolor[RGB]{46,117,182}{\uwave{\textit{珠片绣花}}}，时尚大方，简约 \\
		& \textcolor[RGB]{84,130,52}{\underline{\textit{圆领}}}设计，轻松\textcolor[RGB]{46,117,182}{\uwave{\textit{修饰颈部线条}}}。 \\
		& (Made from \textcolor[RGB]{84,130,52}{\underline{\textit{modal knit fabric}}}. \textcolor[RGB]{46,117,182}{\uwave{\textit{Incredibly soft and stretchy}}}. \textcolor[RGB]{46,117,182}{\uwave{\textit{Slightly looser fit}}}. Flower  print and\\
		&  \textcolor[RGB]{46,117,182}{\uwave{\textit{glitter embroidery}}}. Stylish and casual. The basic \textcolor[RGB]{84,130,52}{\underline{\textit{crewneck}}} design easily helps in \textcolor[RGB]{46,117,182}{\uwave{\textit{elongating the neck}}}. )\\
		\midrule
		\midrule
		\multicolumn{1}{l|}{Input } & 吉普，羽绒服，男，中长款，90绒，冬季外套，新品，连帽，加绒，加厚，保暖羽绒外套， \\
		\multicolumn{1}{l|}{Description} & 黑色，白鸭绒，时尚都市，型男都这么穿，融合艺术细节，创造76年传奇，潮流趋势必备， \\
		& 温暖新升级 ...\\
		& (Jeep's men's down jacket. Mid-thigh length. Filled with 90\% down. Winter jacket. This winter's  \\
		& new arrival. Hoodedd. The down fill provides extra warmth. Warm down jacket. Black. White  \\
		& duck down. Urban style. Built for a perfect look. Designed with artistic details. Creating a legend  \\
		& for 76 years. A must-have to keep up with current fashion trends. Keeps you warmer than ever ... ) \\
		\midrule
		\multicolumn{1}{l|}{Reference} & 采用\textcolor[RGB]{84,130,52}{\underline{\textit{聚酯纤维}}}面料，\textcolor[RGB]{46,117,182}{\uwave{\textit{手感柔软}}}，轻盈且透气性较好，穿在身上干爽舒适。内部以 \textcolor[RGB]{84,130,52}{\underline{\textit{白鸭绒}}}进行\\
		& 填充，充绒量较高，\textcolor[RGB]{46,117,182}{\uwave{\textit{柔软蓬松}}}，更有\textcolor[RGB]{46,117,182}{\uwave{\textit{加厚修身}}}的版型设计，保暖效果较好，为您抵御户外严 \\
		& 寒。 \\
		& (Made from \textcolor[RGB]{84,130,52}{\underline{\textit{polyester}}}. \textcolor[RGB]{46,117,182}{\uwave{\textit{Feel soft}}}, lightweight, and breathable. Keeps you dry and comfortable.  \\
		& Filled primarily with \textcolor[RGB]{84,130,52}{\underline{\textit{white duck down}}}, \textcolor[RGB]{46,117,182}{\uwave{\textit{fluffy and light}}}. Features a \textcolor[RGB]{46,117,182}{\uwave{\textit{thick yet slim-fit}}} design. Keeps you \\
		&  warm in cold climates. )\\
		\midrule
		\multicolumn{1}{l|}{E-PLUG} & 这款羽绒服采用\textcolor[RGB]{84,130,52}{\underline{\textit{中长款}}}的版型设计，修饰你的身材线条，而且还不乏\textcolor[RGB]{46,117,182}{\uwave{\textit{优雅稳重}}}气质。\textcolor[RGB]{84,130,52}{\underline{\textit{连帽}}}的， \\
		& 加持增添青春学院风气息。衣上字母印花的点缀，俏皮又减龄。 \\
		& (This down jacket features a \textcolor[RGB]{84,130,52}{\underline{\textit{mid-thigh length}}}, keeping a stylish silhouette and giving you an \textcolor[RGB]{46,117,182}{\uwave{\textit{elegant}}}\\
		&  \textcolor[RGB]{46,117,182}{\uwave{\textit{ and mature look}}}. The \textcolor[RGB]{84,130,52}{\underline{\textit{hood}}} and letter print on the jacket make you  look younger. ) \\
		\midrule
		\multicolumn{1}{l|}{K-PLUG} & 采用\textcolor[RGB]{84,130,52}{\underline{\textit{聚酯纤维}}}面料制成，\textcolor[RGB]{46,117,182}{\uwave{\textit{手感柔软}}}，\textcolor[RGB]{46,117,182}{\uwave{\textit{亲肤透气}}}。内部以\textcolor[RGB]{84,130,52}{\underline{\textit{白鸭绒}}}填充，蓬松度高，\textcolor[RGB]{46,117,182}{\uwave{\textit{轻盈温暖}}}， \\
		& 更有\textcolor[RGB]{84,130,52}{\underline{\textit{连帽}}}设计，可以锁住人体的热量，为您抵御户外寒冷天气，带来舒适的穿着体验。 \\
		& (Made from \textcolor[RGB]{84,130,52}{\underline{\textit{polyester}}}. \textcolor[RGB]{46,117,182}{\uwave{\textit{Soft and breathable}}}. Filled primarily with \textcolor[RGB]{84,130,52}{\underline{\textit{white duck down}}}, fluffy and light.  \\
		& \textcolor[RGB]{46,117,182}{\uwave{\textit{Lightweight but warm}}}. Features a \textcolor[RGB]{84,130,52}{\underline{\textit{hooded design}}}. Locks in the heat  and keeps out the cold. Comfortable\\
		&  to wear. )\\
		\bottomrule
	\end{tabular}%
		\caption{Case study for the abstractive product summarization task (\textit{Clothing} category). The K-PLUG model generates summaries describing more information about   \textcolor[RGB]{84,130,52}{\underline{\textit{e-commerce knowledge bases}}} and \textcolor[RGB]{46,117,182}{\uwave{\textit{unique selling propositions of product entities}}}. }
	\label{tab:sup_psum_clothing}%
\end{table*}%

\begin{table*}\small
	\centering

	\begin{tabular}{l|l}
		\toprule
		\multicolumn{1}{l|}{Input } & 菲安妮，秋季新款，斜跨包，女，印花磁扣，小方包，时尚单肩包，精致长款肩带，匀 \\
		\multicolumn{1}{l|}{Description} & 整车线工艺，高档ykk拉链，手感丝滑柔软且不脱色，优选进口水珠纹pvc，logo印花与 \\
		& 包身融为一体，手感柔软舒适，防水耐磨，皮质肩带轻便减压，长度可调节，单肩/斜挎 \\
		& 更舒心，平整均匀的车缝线技术 \\
		& (Fion. This fall's new arrival. Corss body bag for women. Flower print magnetic snap closure.  \\
		& Square-shaped. Stylish tote bag. Well made long bag straps. Finished with flat lock stiching.  \\
		& Zippers produced by YKK. Flat and smooth surface. Anti-fading coloring. Made from imported   \\
		&PVC coated fabric. Logo print fits nicely. Soft and comfortable. Water-proof and  durable.\\
		& Lightweight leather shoulder strap. Adjustable length. 2 ways to carry. Finished  with flat and even\\
		&   flatlock stiching.) \\
		\midrule
		\multicolumn{1}{l|}{Reference} & 采用进口水珠纹\textcolor[RGB]{84,130,52}{\underline{\textit{pvc}}}面料制作，\textcolor[RGB]{46,117,182}{\uwave{\textit{手感柔软舒适}}}，\textcolor[RGB]{46,117,182}{\uwave{\textit{防水耐磨耐用}}}。品牌定制拉链，平滑顺畅不 \\
		& 卡链。\textcolor[RGB]{84,130,52}{\underline{\textit{长款}}}肩带，\textcolor[RGB]{46,117,182}{\uwave{\textit{长度可调节}}}，皮质\textcolor[RGB]{46,117,182}{\uwave{\textit{轻便减压}}}，\textcolor[RGB]{84,130,52}{\underline{\textit{单肩}}}\textcolor[RGB]{84,130,52}{\underline{\textit{斜挎}}}随心。 \\
		& (Made from imported \textcolor[RGB]{84,130,52}{\underline{\textit{PVC}}} coated fabric. \textcolor[RGB]{46,117,182}{\uwave{\textit{Soft and comfortable}}}. \textcolor[RGB]{46,117,182}{\uwave{\textit{Water-proof and durable}}}.  Specially\\
		&  made smoothly meshing zippers. \textcolor[RGB]{84,130,52}{\underline{\textit{Long}}} shoulder strap with \textcolor[RGB]{46,117,182}{\uwave{\textit{adjustable length}}}, made with leather for \\
		& \textcolor[RGB]{46,117,182}{\uwave{\textit{light weight and reduced pressure}}}. 2 ways to wear: \textcolor[RGB]{84,130,52}{\underline{\textit{cross body}}}  or\textcolor[RGB]{84,130,52}{\underline{\textit{with the top handles}}}. )\\
		\midrule
		\multicolumn{1}{l|}{E-PLUG} & 这款\textcolor[RGB]{84,130,52}{\underline{\textit{单肩}}}包采用了优质的\textcolor[RGB]{84,130,52}{\underline{\textit{pvc}}}材料制作，它表面具有细腻的纹理，而它的包身上 还具有精美\\
		& 的印花装饰，整体看上去非常优雅，而其内部空间也很大，所以带着它外出非常的方便。 \\
		& (This \textcolor[RGB]{84,130,52}{\underline{\textit{tote bag}}} is made friom imported high-qiuality \textcolor[RGB]{84,130,52}{\underline{\textit{PVC}}} coated fabric. It has  delicate texture and\\
		&  flower print decorations, giving it an elegant look. It can hold many items, making it ideal for every\\
		&   day use.)  \\
		\midrule
		\multicolumn{1}{l|}{K-PLUG} & 优选进口水珠纹\textcolor[RGB]{84,130,52}{\underline{\textit{pvc}}}，logo印花与包身融为一体，\textcolor[RGB]{46,117,182}{\uwave{\textit{手感柔软舒适}}}，\textcolor[RGB]{46,117,182}{\uwave{\textit{防水耐磨}}}，皮质肩带\textcolor[RGB]{46,117,182}{\uwave{\textit{轻便减}}} \\
		& \textcolor[RGB]{46,117,182}{\uwave{\textit{压}}}，\textcolor[RGB]{46,117,182}{\uwave{\textit{长度可调节}}}，\textcolor[RGB]{84,130,52}{\underline{\textit{单肩}}}\textcolor[RGB]{84,130,52}{\underline{\textit{斜挎}}}更舒心，平整均匀的车缝线技术，细节之处彰显品质。 \\
		& (Made from imported \textcolor[RGB]{84,130,52}{\underline{\textit{PVC}}} coated fabric. Logo print fits nicely. \textcolor[RGB]{46,117,182}{\uwave{\textit{Soft and comfortable}}}.  \textcolor[RGB]{46,117,182}{\uwave{\textit{Water-proof and }}}\\
		&\textcolor[RGB]{46,117,182}{\uwave{\textit{durable}}}. Leather shoulder strap for \textcolor[RGB]{46,117,182}{\uwave{\textit{light weight and reduced pressure}}} with \textcolor[RGB]{46,117,182}{\uwave{\textit{adjustable length}}}. 2 ways \\
		&  to  wear: \textcolor[RGB]{84,130,52}{\underline{\textit{cross body }}} or \textcolor[RGB]{84,130,52}{\underline{\textit{with the top handles}}}. Flat  and even flatlocking finish shows its high quality. )\\
		\midrule
		\midrule
		\multicolumn{1}{l|}{Input } & 美旅铝框拉杆箱，男女行李箱，20英寸，登机箱，万向轮，艺术旅行，坚韧抗压，打破时空 \\
		\multicolumn{1}{l|}{Description} & 的边界，突破风格的局限，灵活流畅，双排滑轮，内外兼修干湿分离，防护升级，包角设计 \\
		& ，三重防爆锁扣，美观更牢固，复古亦时尚，且具有良好的密封性 \\
		& (AmericanTourister's aluminum frame suitcase. For men and women. 20 Inch. Carry-on luggage. \\
		&  Spinner wheels. Essential for the creative journey. Strong and durable. Breaks the limits of time  \\
		& and style. Stable and effortless steering thanks to the doubled wheel design. Includes dividers  for easy\\
		&  organization of packed items. Extra stability with corner protection. Triple locks for added  security.\\
		& Stylish design without compromising on durability. Classic yet fashionable. Closes and   seals nicely. ) \\
		\midrule
		\multicolumn{1}{l|}{Reference} & 包角设计，显得更加简洁利落，而且又能有效[防撞耐摔]。\textcolor[RGB]{84,130,52}{\underline{\textit{铝合金}}}框架结构，坚实牢固， 具有\\
		& \textcolor[RGB]{46,117,182}{\uwave{\textit{良好的密封性}}}。\textcolor[RGB]{46,117,182}{\uwave{\textit{三重防爆锁扣}}}更加稳固，让你出门安心无忧。内外间修，\textcolor[RGB]{46,117,182}{\uwave{\textit{干湿分离}}}，合理收纳 \\
		& 更加有序。 \\
		& (Conrer protection gives and basic and simple look with\textcolor[RGB]{46,117,182}{\uwave{\textit{added stability and durability}}}.  \textcolor[RGB]{84,130,52}{\underline{\textit{Aluminum alloy }}}\\
		&  frame makes it strong and \textcolor[RGB]{46,117,182}{\uwave{\textit{close and seal nicely}}}. \textcolor[RGB]{46,117,182}{\uwave{\textit{Triple locks}}}  for added  stability and security. Includes\\
		&  \textcolor[RGB]{46,117,182}{\uwave{\textit{dividers}}} for easy and organized packing.) \\
		\midrule
		\multicolumn{1}{l|}{E-PLUG} & 这款拉杆箱选用干净的\textcolor[RGB]{84,130,52}{\underline{\textit{黑色}}}调，适合多种场合携带。精选材质，具有细腻的纹理质感， \textcolor[RGB]{46,117,182}{\uwave{\textit{经久耐用}}}。\\
		& 内部大空间处理，可以放置更多物品。 \\
		& (This suitcase in \textcolor[RGB]{84,130,52}{\underline{\textit{black}}} is suitable for various occasions. It is made from high-quality materials  with\\
		&  delicate texture and [increased durability]. This large suitcase is built to provide room for extra items.) \\
		\midrule
		\multicolumn{1}{l|}{K-PLUG} & 采用\textcolor[RGB]{84,130,52}{\underline{\textit{铝合金}}}框架结构，坚韧抗压，具有\textcolor[RGB]{46,117,182}{\uwave{\textit{良好的密封性}}}，配备\textcolor[RGB]{46,117,182}{\uwave{\textit{三重防爆锁}}}扣，美观更牢固，搭载内 \\
		& 置式\textcolor[RGB]{46,117,182}{\uwave{\textit{八轮设计}}}，灵活流畅，搭配\textcolor[RGB]{46,117,182}{\uwave{\textit{tsa海关密码锁}}}，有效保护您的物品安全。 \\
		& (Made with \textcolor[RGB]{84,130,52}{\underline{\textit{aluminum alloy}}} frame for added stability. \textcolor[RGB]{46,117,182}{\uwave{\textit{Closes and seals nicely}}}. \textcolor[RGB]{46,117,182}{\uwave{\textit{Triple-lock}}} design to\\
		&  provide extra security. Stable and effortless steering thanks to the 8-wheel design. \textcolor[RGB]{46,117,182}{\uwave{\textit{TAS-approved locks}}}  \\
		&  protect your items and cause no damage during security check.) \\
		\bottomrule
	\end{tabular}%
		\caption{Case study for the abstractive product summarization task (\textit{Bags\&Cases} category). The K-PLUG model generates summaries describing more information about   \textcolor[RGB]{84,130,52}{\underline{\textit{e-commerce knowledge bases}}} and \textcolor[RGB]{46,117,182}{\uwave{\textit{unique selling propositions of product entities}}}. }
	\label{tab:sup_psum_bag}%
\end{table*}%

\begin{table*}\small
	\centering

	\begin{tabular}{l|l}
		\toprule
		\multicolumn{1}{l|}{Q1} & 是纸皮核桃吗 \\
		& (Is it paper-peel walnut?) \\
		\midrule
		\multicolumn{1}{l|}{A1} & 是的哦 \\
		& (Yes, it is.) \\
		\midrule
		\multicolumn{1}{l|}{Q2} & 发的是特级吗 \\
		& (Is it of premium quality?) \\
		\midrule
		\multicolumn{1}{l|}{A2} & 什么意思 \\
		& (What do you mean?) \\
		\midrule
		\multicolumn{1}{l|}{Q3} & 核桃不是有大小吗 我要送人去 买10斤 人在吗 \\
		& (Aren't these walnuts sized differently? I'm getting 5 kilos. Are you still here?)\\
		\midrule
		\multicolumn{1}{l|}{E-PLUG} & 这是 纸皮 核桃 手 捏 就 开 哦 很 好 剥 的 呢 \\
		& (This is paper-peel walnut, which you can peel easily with bare hands.) \\
		\midrule
		\multicolumn{1}{l|}{K-PLUG} & 一般直径在3-5cm \\
		& (They mostly have a diameter between 3-5 cm.)\\
		\midrule
		\midrule
		\multicolumn{1}{l|}{Q1} & 这两个有什么区别 \\
		& (What's the difference between these two?) \\
		\midrule
		\multicolumn{1}{l|}{A1} & 您好上面的纯棉的尺寸1520 \\
		& (Hi, the cotton one above has a size of 1520.) \\
		\midrule
		\multicolumn{1}{l|}{Q2} & 这三种就是大小不一样啊 \\
		& (These three only differ in size?) \\
		\midrule
		\multicolumn{1}{l|}{E-PLUG} & 是的哦亲 以上是3种干巾的区别哦 用途都一样呢 \\
		& (Sure. That's exactly the difference between these three wiping products. They are made  \\
		& for the same purpose.) \\
		\midrule
		\multicolumn{1}{l|}{K-PLUG} & 中间的无纺布的尺寸1820 下面的天然纤维尺寸2020 加大加厚款 \\
		& (The middle one made of nonwoven fabric has a size of 1820, and the bottom one made \\
		& of natural fiber, with a size of 2020, is extra large and thick.) \\
		\midrule
		\midrule
		\multicolumn{1}{l|}{Q1} & 这款茶刀包邮吗。 \\
		& (Do you offer free shipping for this tea knife?)  \\
		\midrule
		\multicolumn{1}{l|}{A1} & 包邮的哦。 \\
		& (Yes we do.)\\
		\midrule
		\multicolumn{1}{l|}{Q2} & 多少钱啊 \\
		& (How much is it?) \\
		\midrule
		\multicolumn{1}{l|}{E-PLUG} & 主要 是 我们 30 箱 之下 就是 这个 价格 了 呢 \\
		& (It's because we have this set price for all orders below 30 boxes.) \\
		\midrule
		\multicolumn{1}{l|}{K-PLUG} & 单把茶刀48哦 \\
		& (It's 48 yuan for a single tea knife.) \\
		\bottomrule
	\end{tabular}%
		\caption{Case study for the multi-turn dialogue task on the ECD dataset. The K-PLUG model produces more accurate responses for the questions related to
		e-commerce knowledge bases.}
	\label{tab:sup_dialogue}%
\end{table*}%

\end{CJK*}
\end{document}